\documentclass{article}
\usepackage[final]{neurips_2024}
\usepackage{footmisc}
\usepackage[utf8]{inputenc} % allow utf-8 input
\usepackage[T1]{fontenc}    % use 8-bit T1 fonts
\usepackage{hyperref}       % hyperlinks
\usepackage{url}            % simple URL typesetting
\usepackage{booktabs}       % professional-quality tables
\usepackage{amsfonts}       % blackboard math symbols
\usepackage{nicefrac}       % compact symbols for 1/2, etc.
\usepackage{microtype}      % microtypography
\usepackage{xcolor}         % colors
\usepackage{graphicx}
\usepackage{amsmath}
\usepackage{amssymb}

\newcommand{\MyEmoji}[1]{\includegraphics[width=1em,valign=t]{#1}}

\newcommand{\gear}{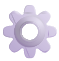}
\usepackage[export]{adjustbox}
\usepackage{float}
\usepackage{tabularray}
\UseTblrLibrary{booktabs}
\usepackage{tabularx}
\usepackage{booktabs}
\usepackage{multirow}
\usepackage{tabularray}

\usepackage{marvosym}
\usepackage{tikz}
\usetikzlibrary{spy}
\usetikzlibrary{shapes.misc}
\usepackage{hyperref}
\usepackage{cleveref}
\crefname{figure}{Fig.}{Fig.}
\crefname{table}{Tab.}{Tab.}
\crefname{equation}{Eq.}{Eq.}
\crefname{section}{Sec.}{Sec.}

\usepackage{animate}
\usepackage{graphicx}

\newcommand{\ie}{\textit{i}.\textit{e}.}
\newcommand{\eg}{\textit{e}.\textit{g}.}

\newcommand\nnfootnote[1]{%
  \begin{NoHyper}
  \renewcommand\thefootnote{}\footnote{#1}%
  \addtocounter{footnote}{-1}%
  \end{NoHyper}
}
\renewcommand{\thefootnote}{}
\hypersetup{
    colorlinks=true,
    linkcolor=black,
    citecolor=black,
    urlcolor=purple % 设置 URL 颜色为深紫色
}

\begin{document}

\title{SpikeReveal: Unlocking Temporal Sequences from Real Blurry Inputs with Spike Streams}

\author{
Kang Chen$^{1,2*}$ \quad Shiyan Chen$^{1,2*}$ \quad Jiyuan Zhang$^{1,2}$ \quad Baoyue Zhang$^{1,2}$ \and \textbf{Yajing Zheng}$^{1,2}$\textsuperscript{\Letter} \quad \textbf{Tiejun Huang}$^{1,2,3}$ \quad \textbf{Zhaofei Yu}$^{1,2,3}$\textsuperscript{\Letter}  \\
$^1$ School of Computer Science, Peking University \\
$^2$ National Key Laboratory for Multimedia Information Processing, Peking University\\
$^3$ Institute for Artificial Intelligence, Peking University \\
\texttt{\{mrchenkang,strerichia002p,jyzhang,byzhang\}@stu.pku.edu.cn} \\
\texttt{\{yj.zheng,tjhuang,yuzf12\}@pku.edu.cn}
}

\maketitle

\begin{abstract}
Reconstructing a sequence of sharp images from the blurry input is crucial for enhancing our insights into the captured scene and poses a significant challenge due to the limited temporal features embedded in the blurry image. Spike cameras, sampling at rates up to 40,000 Hz, have proven effective in capturing motion features and beneficial for solving this ill-posed problem. 
Nonetheless, existing methods fall into the supervised learning paradigm, which suffers from notable performance degradation when applied to real-world scenarios that diverge from the synthetic training data domain.
% todo 
% Moreover, the quality of reconstructed images is capped by the generated images based on motion analysis interpolation, which inherently differs from the actual scene, affecting the generalization ability of these methods in real high-speed scenarios. 
To address this challenge, we propose the first self-supervised framework for the task of spike-guided motion deblurring. Our approach begins with the formulation of a spike-guided deblurring model that explores the theoretical relationships among spike streams, blurry images, and their corresponding sharp sequences. We subsequently develop a self-supervised cascaded framework to alleviate the issues of spike noise and spatial-resolution mismatching encountered in the deblurring model.
With knowledge distillation and reblur loss, we further design a lightweight deblur network to restore high-quality sequences with brightness and texture consistency with the original input. Quantitative and qualitative experiments conducted on our real-world and synthetic datasets with spikes validate the superior generalization of the proposed framework. Our code, data and trained models are available at \url{https://github.com/chenkang455/S-SDM}.

\end{abstract}
\begin{figure}[t]
    \centering
    \includegraphics[width=1.0\linewidth]{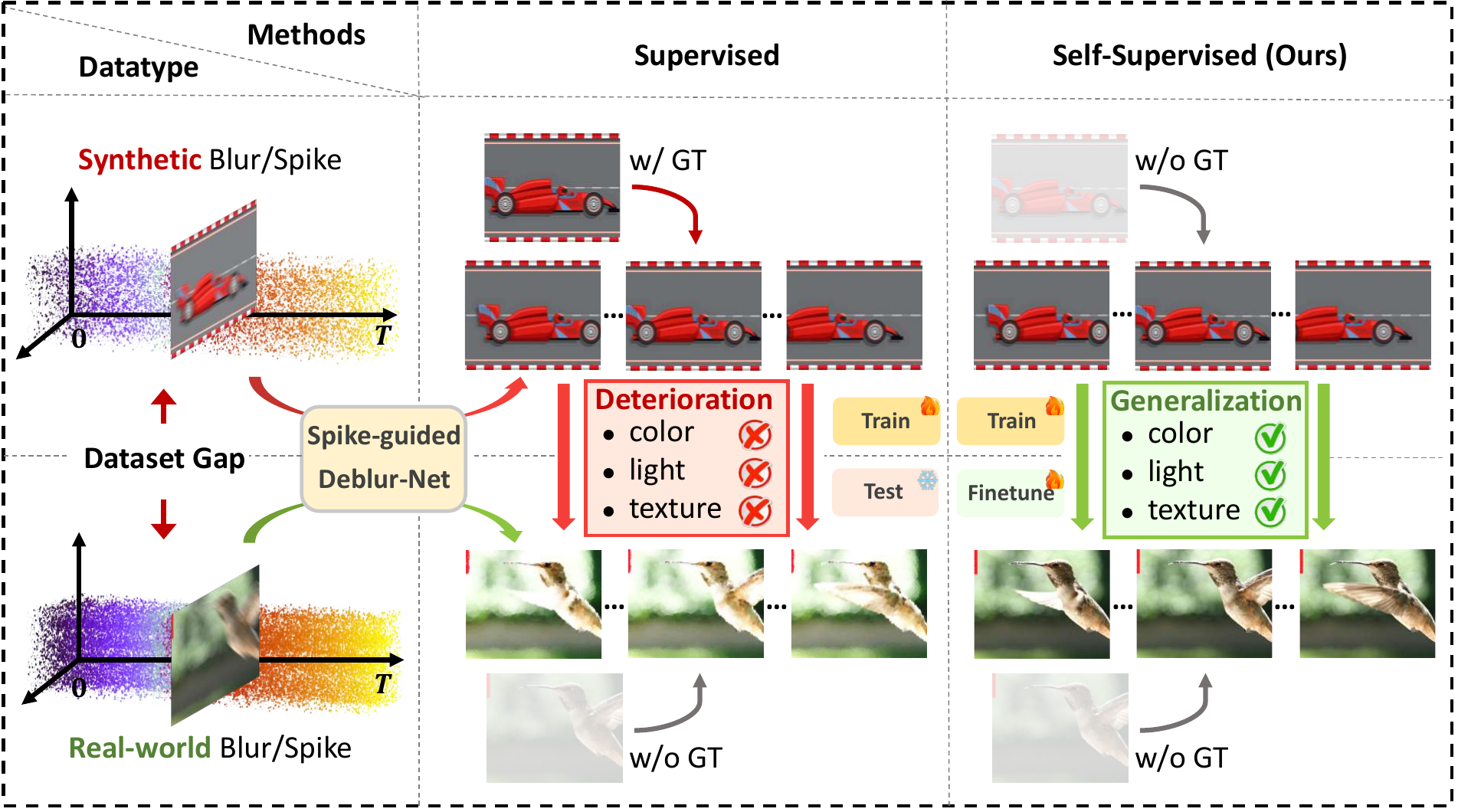}
    \caption{Illustration of the superiority of our self-supervised framework (S-SDM) over supervised methods. Supervised methods, while effective on synthetic datasets, suffer from a significant performance decline when applied to real-world datasets, primarily due to data distribution discrepancies. In contrast, our self-supervised framework, necessitating no Ground Truth (GT) for training, seamlessly bridges this dataset gap through fine-tuning on real-world datasets.}
    \label{fig:fig1}
    \vspace{-1em}
\end{figure}

\nnfootnote{\textsuperscript{$*$} Equal contributors.}
\nnfootnote{\textsuperscript{\Letter} Corresponding authors.}

\section{Introduction}
\label{sec:intro}
Traditional cameras, constrained by their exposure-based imaging mechanism, often produce blurry images when capturing fast-moving objects or during camera movement throughout the exposure process~\cite{exposure,Survey}.
% While these blurry images lose significant details, the ability to recover dynamic motion trajectories from the static blurry input would greatly enhance our capacity to extract valuable information about the captured scene. 
While these blurry images lose significant details, the ability to recover dynamic motion trajectories from the static blurry input becomes critically important. 
% todo These blurry images, while losing significant detail, present an opportunity to extract valuable information about the scene by recovering dynamic motion trajectories from static blurry inputs.
%However, due to the limited motion features embedded in blurry frames, the situation of multiple motion trajectories corresponding to the same blurry input may exist \cite{E-CIR,EFNet,evdi,gem,red}, such as two objects moving along the same trajectory but in opposite directions, which introduces ambiguity into the task of motion deblurring. Recently, learning-based approaches \cite{motion_etr,levs,bit} aim to address this issue by directly establishing mappings from blurry input to sharp sequences based on datasets. Nevertheless, traditional cameras still have limited capabilities in capturing high-speed motion details during exposure, resulting in performance degradation of these methods in scenarios beyond the dataset.
However, the inherent challenge lies in the limited motion features available within blurry frames, leading to potential ambiguities such as multiple motion trajectories corresponding to the same blurry input. This is exemplified by scenarios where two objects move along the same trajectory but in opposite directions \cite{E-CIR,EFNet,evdi}, rendering the task of motion deblurring ill-posed. Recent advancements in learning-based approaches \cite{motion_etr,levs,bit} seek to address this challenge by establishing direct mappings from blurry inputs to sharp sequences in a supervised learning manner. Despite these efforts, traditional cameras struggle to capture fine details in high-speed motion due to their exposure constraints, thus limiting the effectiveness of these methods in scenarios not covered by the training datasets.

In recent years, neuromorphic cameras \cite{event_review,spikecamera}, leveraging their ultra-high temporal resolution and high dynamic range, have found widespread use in many fields, including computer vision and robotics. These cameras, including event and spike cameras, are distinguished by their ability to produce high temporal resolution outputs directly tied to changes in light intensity. Specifically, event cameras generate events in areas where light intensity changes \cite{trmd}, while spike cameras capture the absolute brightness of the scene at each pixel, offering a stream of spikes as output \cite{spikecamera}.
%Neuromorphic cameras primarily consist of event and spike cameras, both of which output ultra-high temporal resolution events or spike streams directly based on changes in light intensity. The distinction lies in that event streams are generated in regions of motion \cite{trmd}. In contrast, spike streams, which represent the absolute brightness of the scene, are produced at each pixel position \cite{spikecamera}. 
This distinctive feature endows spike cameras with a significant advantage \cite{ tfp_tfi,STDP_zheng,zhu2020retina,bsn_chen,spk2img,red_chen} in capturing and recovering sharp texture from scenes with rapid motion. 
% However, spike cameras essentially provide single-channel pixel variations, meaning grayscale images are obtained solely based on spike streams. Additionally, the output spatial resolution of ultra-high-speed neural morphological cameras is not very high.

%todo 在这段话中要借助头图描述
%The challenge addressed in this work is how to leverage the strengths of traditional and spike cameras to recover a sharp sequence from the blurry input in a self-supervised manner, as shown in \cref{fig:fig1}. 
% This study tackles the challenge of harnessing the strengths of both traditional and spike cameras to reconstruct sharp sequences from blurry inputs in a self-supervised manner, as depicted in \cref{fig:fig1}.
% While recent studies \cite{1000FPS,spike_deblur} have explored the potential of RGB-Spike fusion, these efforts predominantly operate within the supervised learning framework,
%their frameworks are constrained within the supervised learning paradigm, 
Recent studies \cite{1000FPS,spike_deblur} have explored the potential of RGB-Spike fusion, \ie, harnessing the strengths of both traditional and spike cameras to reconstruct sharp sequences from blurry inputs. However, their frameworks are constrained within the supervised learning paradigm, which necessitates extensive datasets comprising pairs of blurry and sharp images, as well as spike sequences.
While synthetically acquiring such paired data, as demonstrated in previous studies \cite{spike_deblur,trmd,EFNet}, is feasible, collecting them in real-world scenarios presents the following challenges: (1) high-speed cameras are prohibitively expensive and not readily deployable in many settings; (2) spatial-temporal calibration between spike cameras and high-speed RGB cameras complicates the data collection process. 
These problems render the fine-tuning of supervised methods on real-world datasets challenging, further leading to their performance deterioration in such environments as shown in \cref{fig:fig1}. The resulting degradation in image quality, manifesting as color distortion, brightness inconsistency, and inaccurate texture restoration, is mainly caused by the disparity between synthetic and real-world datasets, especially in terms of the density of spike stream, spike generation mechanism, and blurry image generation.
Moreover, the effectiveness of supervised methods is inherently limited by the ground truth sequences created through motion analysis interpolation 
algorithms \cite{sim2021xvfi,RIFE}, which inherently differs from the real-world scene and thus affects the model's generalization ability.

To overcome these issues, we propose the first-of-its-kind \textbf{S}elf-supervised \textbf{S}pike-guided \textbf{D}eblurring \textbf{M}odel \textbf{(S-SDM)}, capable of recovering the continuous sharp sequence from a single blurry input with the assistance of low-resolution spike streams.
We begin with a theoretical analysis of the relationship between spike streams, blurry images, and sharp sequences, leading to the development of our Spike-guided Deblurring Model (SDM). We further construct a self-supervised processing pipeline by cascading the denoising network and the super-resolution network to reduce the sensitivity of the SDM to spike noise and its reliance on spatial-resolution matching between the two modalities. To reduce the computational cost and enhance the utilization of spatial-temporal spike information within this pipeline, we further design a Lightweight Deblurring Network (LDN) and train it based on pseudo-labels from the teacher model, \ie, the established self-supervised processing pipeline. Further introducing reblur loss during LDN training, we achieve better restoration performance and faster processing speed than the processing-lengthy and structure-complicated teacher model. To validate the performance of our S-SDM across various scenarios, we build an RGB-Spike binocular system and propose the first spatially-temporally calibrated Real-world Spike Blur (RSB) dataset in this community. Quantitative and qualitative experiments conducted on the real-world and synthetic datasets validate the superiority of our method. In summary, our key contributions are:
\begin{itemize}
% \vspace{-0.4em}
\setlength{\itemsep}{0pt}
\setlength{\parsep}{0pt}
\setlength{\parskip}{0pt}
\item[$\bullet$] We develop a self-supervised spike-guided image deblurring framework, addressing the performance degradation due to the synthetic-real domain gap in supervised methods.
\item[$\bullet$] We perform an in-depth theoretical analysis of the fusion between the spike stream and blurry image, leading to the development of the SDM.
% \item[$\bullet$] We propose a self-supervised deblurring model that leverages distillation learning, enabling the training of the LDN in real-world scenarios.
\item[$\bullet$] We propose a real-world dataset RSB and experiments on GOPRO and RSB  datasets validate the superior generalization of our S-SDM.
% \vspace{-0.8em}
\end{itemize}

\section{Related Work} 
\noindent\textbf{Spike Camera.} The spike camera, inspired by the primate retina, stands apart from conventional cameras with its ability to generate synchronous spike streams for each pixel at extremely low latency. This distinct feature provides significant advantages in various applications such as high-speed imaging \cite{tfp_tfi,STDP_zheng,zhu2020retina,bsn_chen,spk2img,red_chen,wgse}, optical flow estimation \cite{spike_flow}, object detection \cite{motion_estimation_zheng}, 3D reconstruction \cite{zhang2024spikegs}, depth estimation \cite{spikedepth}, motion deblurring \cite{spike_deblur,zhang2024deblur}, and occlusion removal \cite{spike_sai}. 
% In the realm of high-speed image reconstruction,  \cite{tfp_tfi} initially proposed the texture from play-back (TFP) image reconstruction method based on the spike's generation mechanism and virtual exposure imaging techniques. 
% Inspired by the spike camera's biological principles, studies like  \cite{STDP_zheng,zhu2020retina} have employed short-term synaptic plasticity and retinal imaging principles to transform the spike stream into the high frame rate video sequence. Zhao et al.~\cite{zhao2022spikingsim} proposed the first end-to-end CNN-based model to promote the quality of spike-based high-speed imaging.
%However, these approaches often suffer from significant image quality degradation in real-world scenarios due to inadequate modeling of spike noise. Addressing this,  \cite{spk2img} leveraged the powerful nonlinear fitting capabilities of CNNs to train an end-to-end model for converting the spike stream into sharp images on synthetic datasets. 
% To overcome the performance degradation when applying methods trained on the synthetic dataset to real-world scenarios, \cite{red_chen} constructed a joint optimization framework for the spike reconstruction network and the optical flow estimation network, enabling fine-tuning for real-world scenarios.

\noindent\textbf{Spike-guided Motion Deblurring.} 
While the spike camera boasts an ultra-high temporal resolution, its development is currently impeded by the low spatial resolution. 
% arising from the limitation in data transmission bandwidth and manufacturing processes. 
Additionally, the single-channel output from the spike camera restricts previous methods from recovering the image color information. 
To address these issues, a promising approach is establishing an RGB-Spike hybrid imaging system  \cite{1000FPS}. The binocular system achieves the multi-modality fusion of High-spatial/Low-temporal RGB blurry input and High-temporal/Low-spatial spike stream, thereby also serving as a spike-guided motion deblurring method  \cite{spike_deblur}. 
However, to the best of our knowledge, existing spike-guided deblurring methods  \cite{1000FPS,spike_deblur} predominantly rely on supervised training on synthetic datasets. This reliance results in significant performance degradation when these methods are evaluated in real-world scenarios due to the domain discrepancies between synthetic and real datasets as illustrated in \cref{fig:fig1}. 
% Addressing this issue, we propose a novel physical model and a self-supervised framework specifically for the spike-guided motion deblurring task.

% event-based deblur related-work 需要加上吗？

\noindent\textbf{Event-based Motion Deblurring.} Event camera  \cite{eventcamera1} can asynchronously generate events that record log-intensity changes at the pixel level with minimal latency, which contains a rich set of motion features beneficial for motion deblurring tasks. 
% A notable contribution in this domain was made by  \cite{EDI}, which formulated the physical model for event-based motion deblurring as a single-variable non-convex optimization problem. However, the deblurred sequence produced by this model lacks the expected sharpness due to the reliance of a fixed trigger threshold and the influence of noisy events.
Numerous supervised methods  \cite{LEDIVDI,trmd,RES, LEDIVDI, E-CIR,EFNet} have been proposed to learn the mapping from the blurry input, events to the sharp outcome. Despite these advancements, a major hurdle remains in obtaining real blurry-sharp image pairs for training. 
% While the aforementioned methods trained networks on synthetic datasets and evaluated them on real-world datasets, they suffer from significant performance drops due to the discrepancies between synthetic and real-world datasets.
To overcome the domain gap between synthetic and real-world datasets, recent methods \cite{red,evdi,gem} explored the mutual constraint between the blurry image and event stream, enabling the training of networks on real-world blur datasets.
% , offering a practical solution to this longstanding challenge.\cite{red} introduced a method that enforces a constraint between the outputs of the deblurring and optical networks, enabling the fine-tuning of the deblur network on real data.  \cite{evdi} tackled motion deblurring and interpolation tasks simultaneously by leveraging the relationship between adjacent blurry images and events. Additionally,  \cite{gem} developed a scale-aware network capable of handling flexible input spatial scales. This approach facilitates learning from various temporal scales of motion blur, thereby enhancing the performance of event-based deblurring in real-world situations.

% 方法部分的细节reading
\section{Method}
\subsection{Preliminaries}
\noindent\textbf{Spike Camera Mechanism.} Consider $\mathbf{L}(t)$ to represent the latent sharp frame at time $t$. Each pixel $p$ in the spike camera \cite{spikecamera} has an integrator that accumulates the incoming photons at a high frequency. Once the cumulative intensity exceeds a predefined threshold $C$ at time $t_e$, pixel $p$ emits a spike, and the accumulation of photons is reset to zero. This process can be mathematically described as follows:
\begin{equation}
\int_{t_s}^{t_e}\mathbf{L}(t)dt \geq C, \label{equ:spike_generation}
\end{equation}
where $t_s$ denotes the firing time of the previous spike. While the spike camera is capable of generating asynchronous spike streams akin to that of event cameras,  its effectiveness is constrained by the inherent limitations of its physical circuitry, which necessitates reading spikes at a predetermined sampling rate. We denote the generated spike stream as $\mathcal{S} \in \{0,1\}^{K\times 1\times H\times W}$, where $H$ and $W$ signify the height and width of the image, and $K$ represents the length of the spike sequence. 
% $\mathcal{S}_{p}[i] = 1$ indicates that pixel $p$ generates a spike in the $i$-th frame, and the inverse implies the absence of a spike.

\noindent\textbf{Problem Formulation.} In traditional photography, motion blur occurs when there is relative movement between the camera and the scene during the exposure period. According to the motion blur physical model  \cite{event_review}, the blurry image $\mathbf{B}$ can be represented as the average of the latent frame $\mathbf{L}(t)$ over the exposure $\mathcal{T}$, \ie:
\begin{equation}
\mathbf{B} = \frac{1}{T}\int_{t\in \mathcal{T}} \mathbf{L}(t) dt,\label{equ:blur_equation}
\end{equation}
where $T$ represents the exposure period. Despite the spike camera's superior temporal resolution, its spatial resolution remains comparatively low.
%While the spike camera boasts an ultra-high temporal resolution, its spatial resolution is still relatively low, mainly due 
This limitation is primarily attributed to the constraints in data transmission bandwidth and the challenges inherent in the manufacturing process.
%to data transmission bandwidth limitations and the manufacturing process's constraints. 
Here, we postulate that the spatial resolution of the spike camera is approximately one-quarter that of a conventional RGB camera.

In this paper, we aim to enhance the High-spatial/Low-temporal resolution blurry input $\mathbf{B} \in \mathbb{R}^{1\times 3\times H \times W}$ into a sequence of High-Quality images $\{\mathbf{L}(t_i)\}_{i=1}^K \in \mathbb{R}^{K\times 3\times H \times W}$ with the aiding of High-temporal/Low-spatial resolution spike stream $\mathcal{S}_{\mathcal{T}} \in \{0,1\}^{K \times1\times \frac{H}{4}\times \frac{W}{4}}$, which can be mathematically formulated as:
\begin{equation}
\label{eq:3}
\{\mathbf{L}(t_i)\}_{i=1}^K = \text{Deblur}(t_i;\mathbf{B},\mathcal{S}_{\mathcal{T}}).
\end{equation}

In \cref{eq:3}, $\text{Deblur}(\cdot)$ represents the Spike-guided
Deblur-Net as shown in \cref{fig:fig1}, $i$ refers to the $i$-th frame in the spike stream $\mathcal{S}_{\mathcal{T}}$, and $t_i$ is the timestamp associated with this frame.

\subsection{Theoretical Analysis} \label{sec:SDM}
The spike camera, with its photodetector tailored to capture the single-channel light intensity, faces difficulties in obtaining the color information that the multi-channel RGB camera can effortlessly capture. Therefore, we modify the color intensity $\mathbf{L}(t)$ in  \cref{equ:spike_generation} to the grayscale value $\mathbf{L}_g(t)$ for further analysis:
\begin{equation}
\int_{t_s}^{t_e}\mathbf{L}_g(t)dt \geq C.\label{equ:spike_generation_gray}
\end{equation}
In this formulation, $\mathbf{L}_g(t) = w_r\cdot \mathbf{L}_r(t)+w_{gre}\cdot \mathbf{L}_{gre}(t)+w_b\cdot \mathbf{L}_b(t)$, where $\mathbf{L}_c(t),w_c$ denote the intensity and weight of channel $c \in\{r,gre,b\}$ respectively.

Given the blurry input $\mathbf{B}$ and its corresponding
spike stream $\mathcal{S}_{\mathcal{T}}$, we incorporate \cref{equ:spike_generation_gray} 
 into the motion blur model presented in \cref{equ:blur_equation}. This integration formulates a link between the two modalities as outlined below:
\begin{equation}
\mathbf{B}_g = \frac{C \cdot N_{\mathcal{T}}}{T}, \label{equ:Blur_Spike_long}
\end{equation}
where $\mathbf{B}_g$ denotes the grayscale version of the blurry input, and $N_{\mathcal{T}}$ denotes the total number of spikes accumulated over the exposure period.
The accumulation $N_{\mathcal{T}}$ is calculated as $N_{\mathcal{T}} = \sum_{i=1}^K \mathcal{S}[i]$, with $\mathcal{S}[i]$ indicating the $i$-th frame of the spike stream.

Within the exposure $\mathcal{T}$, we consider a shorter spike sequence centered around the $t$ moment $\mathcal{S}_{\mathcal{T'}} \in \{0,1\}^{K'\times 1\times H\times W}$, satisfying $K' \ll K$, $t \in \mathcal{T}$, and $\mathcal{T'} \subset \mathcal{T}$. 
Similar to \cref{equ:Blur_Spike_long}, we can derive the relationship between the short-exposure gray image $\mathbf{E}_g(t,\mathcal{T}')$ and the short spike stream $\mathcal{S}_{\mathcal{T'}}$ as follows:
\begin{equation}
\mathbf{E}_g(t,\mathcal{T}') = \frac{1}{T'} \int_{s\in\mathcal{T}'}\mathbf{L}_g(s) ds = \frac{C \cdot N_{\mathcal{T}'}}{T'},\label{equ:Blur_Spike_short}
\end{equation}
where $T' \ll T$ represents the short exposure period. 
% It should be noted that $\textbf{B}_{g}$ represents the gray blurry input, while $B_g(t, \mathcal{T}')$ denotes the gray short-exposure image as illustrated in \cref{fig:SDM_BSN_SR}. 
% We adopt a notation in which the bolded symbol $\mathbf{B}_{g}$ signifies the image captured by conventional cameras, whereas the non-bolded expression $B_g(t, \mathcal{T}')$ corresponds to images reconstructed from the spike stream, as depicted in \cref{fig:SDM_BSN_SR}(a).

Given the observation that the color information of adjacent pixels often exhibits similarity under the premise of minor motion amplitude, we postulate that the colors of the blurry input $\mathbf{B}$ and the short-exposure image $\mathbf{E}(t,\mathcal{T}')$ are identical. This assumption implies that the intensity proportion among RGB channels in the blurry input  $\alpha_c^{\mathbf{B}}$ and the short-exposure image $\alpha_c^{\mathbf{E}}(t,\mathcal{T}')$ is approximately equivalent, satisfying $\alpha_c^{\mathbf{B}} = \mathbf{B}_g / \mathbf{B}_c$ and $\alpha_c^{\mathbf{E}}(t,\mathcal{T}') = \mathbf{E}_{g}(t,\mathcal{T}') /  \mathbf{E}_{c}(t,\mathcal{T}')$. More details can be found in the supplementary materials. 

Upon establishing it, we move forward to build a mathematical relation between the blurry image and the spike stream. By substituting the gray channel $g$ with color channel $c$ and dividing \cref{equ:Blur_Spike_long} by \cref{equ:Blur_Spike_short}, we efficiently eliminate the unknown threshold $C$ and weights $\alpha_c^{\mathbf{B}}/\alpha_c^{\mathbf{E}}(t,\mathcal{T}')$, leading to the following equation:
\begin{equation}
\mathbf{E}_c(t,\mathcal{T}') = \mathbf{B}_c \cdot \frac{N_{\mathcal{T}'}}{N_{\mathcal{T}}} \cdot \frac{T}{T'}.
\label{equ:SDM}
\end{equation}

By applying \cref{equ:SDM} to RGB three channels, we explicitly establish the relationship between the color blurry input $\mathbf{B}$, the color short-exposure image $\mathbf{E}(t,\mathcal{T}')$ and the spike stream $\mathcal{S}_{\mathcal{T}}$ as shown in \cref{sup_fig:s_sdm}. Since the exposure period $\mathcal{T'}$ is relatively short, it is reasonable to assume that the scene remains static. In this context,  we interpret the short-exposure image $\mathbf{E}(t,\mathcal{T}')$ as the latent sharp frame $\mathbf{L}(t)$, allowing us to modify \cref{equ:SDM} as follows:
\begin{equation}
\mathbf{L}(t) = \mathbf{B} \cdot \frac{N_{\mathcal{T}'}}{N_{\mathcal{T}}} \cdot \frac{T}{T'}.
\label{equ:SDM_final}
\end{equation}

To this end, we have conducted a comprehensive theoretical analysis of the spike-guided motion deblurring task, which is neglected in prior learning-based motion deblur methodologies  \cite{1000FPS,spike_deblur}. For further discussion readability, we refer to \cref{equ:SDM_final} as the \textbf{Spike-guided Deblurring Model (SDM)}, which is analogous to the baseline motion deblur model EDI \cite{EDI} in event camera.

% Compared to the learning-based method SpkDeblurNet \cite{spike_deblur}, which relies on training with a large amount of paired data and cumbersome deep networks, our proposed non-learning model SDM achieves multimodal information fusion in a relatively direct manner. 
% \input{figs/main/SDM_BSN_SR}

\begin{figure}[t]
    \centering
    \includegraphics[width=1.0\linewidth]{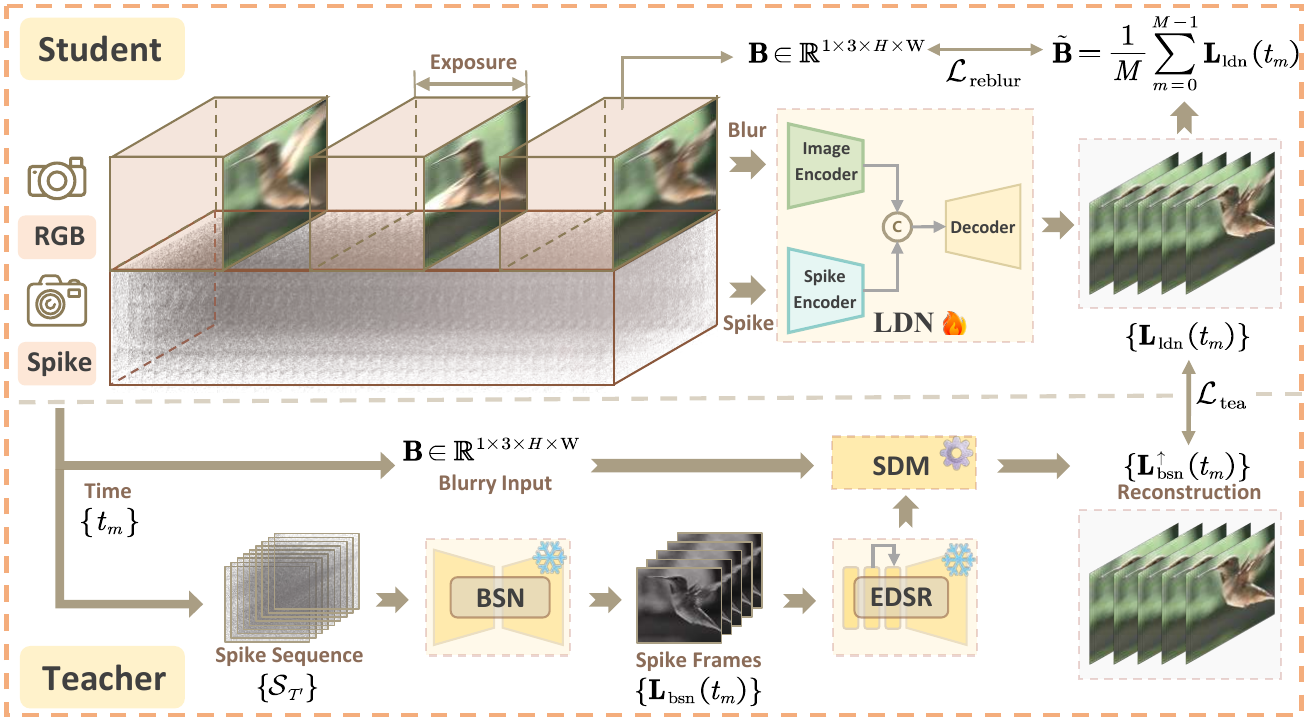}
    \caption{The schematic diagram of our proposed distillation self-supervised framework. The  ``\MyEmoji{\gear}" indicates that certain computations are executed in a non-network manner.}
    \label{fig:distillation}
        \vspace{-1em}
\end{figure}

\subsection{Self-supervised Spike-guided Deblurring Model}\label{sec:self_supervised}
\subsubsection{Processing Pipeline}
While SDM theoretically allows for the fusion of the blurry image and the spike stream, its practical deployment faces the following obstacles: 
\begin{itemize}
% \vspace{-0.4em}
\setlength{\itemsep}{0pt}
\setlength{\parsep}{0pt}
\setlength{\parskip}{0pt}
    \item[$\bullet$] The deblurred image $\mathbf{E}(t,\mathcal{T}')$ suffers from noise-related degradation due to the lack of adequate spike information during the short exposure $\mathcal{T}'$.
    \item[$\bullet$] The spatial resolution of the spike camera is approximately one-quarter of the RGB camera, rendering the SDM implementation impractical.
% \vspace{-0.4em}
\end{itemize}
To overcome these limitations, we further cascade the self-supervised denoising network to eliminate the spike noise in $N_{\mathcal{T}'}$ and super-resolution network to match spatial resolutions of the blurry image and spike stream, with the processing pipeline illustrated in the bottom of \cref{fig:distillation}.

% To tackle these limitations, we devise a lengthy yet feasible processing pipeline based on the SDM as shown in the bottom of \cref{fig:distillation}. Initially, we utilize \cref{equ:Blur_Spike_short} to convert the short-exposure spike stream into an image, followed by a sequential cascade of the Denoising Network and the Super-Resolution Network, which are tasked with denoising and resolution-enhancement respectively in a self-supervised manner. Subsequently, the deblurred image is obtained through the SDM, with inputs comprising both the blurry image and the enhanced short-exposure image. In the following, we delve into the details of the Denoising Network and Super-Resolution Network.

\textbf{Denoising Network.} 
% Recently, deep learning-based supervised denoising methods have seen considerable advancements, achieving superior performance in noise reduction tasks. However, collecting clean and noisy image pairs in this task is challenging.
% requires spatial-temporal calibration between the high-speed camera and the spike camera, which is both time-consuming and expensive. 
We leverage the Blind Spot Network (BSN)  \cite{BSN,bsn_add_1,bsn_add_2,bsn_chen,bsn_add_3,atbsn} to predict the clean spike accumulation $N_{\mathcal{T}'}$ from the input short-exposure spike stream $\mathcal{S}_{\mathcal{T}'}$. 
The core idea of BSN is to design the blind-spot strategy that compels the convolutional layer to estimate the clean value of each pixel solely based on its surrounding pixels. 

Under the premise that the spike stochastic thermal noise is independent identically distributed \cite{zhao2022spikingsim}, the BSN is trained to deduce sharp spike frames from the input, with the loss function formulated as:
\begin{equation}
\mathcal{L}_{\text{BSN}} = ||  \text{BSN}(\mathcal{S}_{\mathcal{T}'};\Theta_{\text{1}}) - N_{\mathcal{T}'})||_2^2 ,
\end{equation}
where the denoised spike frame $\text{BSN}(\mathcal{S}_{\mathcal{T}'};\Theta_{\text{1}})$ is denoted by $\mathbf{L}_{\text{bsn}}(t)$ for further analysis.

% 支撑材料
% We construct our denoising network based on the blind-spot strategy outlined in  \cite{BSN}. We first rotate the input image four times, then concatenate them into the U-Net \cite{UNet} structure like denoising network. To prevent direct mapping from the input pixel to the output pixel, a single-pixel offset strategy is employed in the convolutional kernel to separate its receptive field from the central pixel. Finally, the denoising output is obtained by merging the results of four branches via a $1\times1$ convolution.

\textbf{Super-Resolution Network.} 
% Akin to the denoising network, the field of supervised learning-based Super-Resolution (SR) networks has achieved remarkable progress recently. However, gathering paired High/Low resolution short-exposure images $B(t,\mathcal{T}')$ is challenging due to the scarcity of the high-resolution spike camera. 
In this task, we observe that the blurry input $\mathbf{B}_g$ and the long-exposure spike frame $N_{\mathcal{T}}$ exhibit the same texture features as shown in \cref{equ:Blur_Spike_long}. This observation motivates us to train the Super-Resolution (SR) network based on pairs of the blurry images and the long-exposure spike frames.

We leverage the well-explored Enhanced Deep Super-Resolution network (EDSR) \cite{edsr} as the backbone of our SR network, with the loss function formulated as follows:
\begin{equation}
\mathcal{L}_{\text{EDSR}} = ||\text{EDSR}(N_{\mathcal{T}};\Theta_2) - \mathbf{B}_g||_2^2.
\end{equation}
With the training of the SR network completed, we freeze its parameters and apply it to the denoised spike frame $\mathbf{L}_{\text{bsn}}(t)$, yielding the resolution-enhanced spike frame $\mathbf{L}_{\text{bsn}}^{\uparrow}(t)$.

\subsubsection{Knowledge Distillation  Framework}
While the aforementioned processing pipeline achieves the multi-modality fusion of the  blurry input and the spike stream, several aspects still need refinement:
\begin{itemize}
% \vspace{-0.8em}
\setlength{\itemsep}{0pt}
\setlength{\parsep}{0pt}
\setlength{\parskip}{0pt}
\item[$\bullet$] The framework is lengthy and computationally demanding, which hinders its suitability for real-time system deployment.
\item[$\bullet$] The blind-spot strategy of the BSN limits the full utilization of the spatial information inherent in the spike stream.
\item[$\bullet$] The representation of the short-exposure image does not fully reflect the advantages of the high temporal resolution inherent in the spike stream.
\end{itemize}

To improve them, we further build a knowledge distillation framework building upon the  existing  processing pipeline. This pipeline serves as the teacher model, providing the reconstructed sequence as pseudo-labels for the training of the student model LDN, as illustrated in \cref{fig:distillation}.

\textbf{Lightweight Deblur Network.} LDN adheres to a similar input and output pattern as previous research \cite{spike_deblur}, \ie, taking the blurry input $\mathbf{B}$ and the short spike stream $\mathcal{S}_{\mathcal{T}'}$ centered around moment $t$ as inputs, with the output being the reconstructed sharp image $\mathbf{L}_{\text{ldn}}(t)$, mathematically formulated as follows:
\begin{equation}
\mathbf{L}_{\text{ldn}}(t) = \text{LDN}(\mathbf{B},\mathcal{S}_{\mathcal{T}'};\Theta_3),
\end{equation}
Full details regarding the LDN structure are available in the supplementary materials.
% 补充材料
% where the image encoder consists of two layers of down-sampling convolutions to align the spatial resolution of two modalities. In contrast to the intricate cross-attention mechanism outlined in  \cite{trmd,spike_deblur}, we implement the fusion of two modalities by the simple operation of $\text{Concat}(\cdot)$, with the cascaded CBAM \cite{CBAM} and residual blocks for further feature fusion. 

% In real-world scenarios where ground truth images are unavailable for supervision, we employ the previously described processing pipeline as the teacher model. 
To avoid the scenario where the LDN exactly replicates the mapping of the teacher model, we design the teacher loss $\mathcal{L}_{\text{tea}}$ based on the LPIPS  \cite{lpips} loss and further introduce the blur reconstruction loss. The reblur loss $\mathcal{L}_{\text{reblur}}$ measures the difference between the blurry input $\mathbf{B}$ and the re-synthesized blurry image ${\widetilde{\mathbf{B}}}$, satisfying:
\begin{equation}
 \widetilde{\mathbf{B}}=\frac{1}{M}\displaystyle\sum_{m=1}^{M}\mathbf{L}_{\text{ldn}}(t_m),   
\end{equation}
where $\mathbf{L}_{\text{ldn}}(t_m)$ represents the $m$-th recovered image within the exposure period $\mathcal{T}$ and $M$ is the total number of reconstructed images. Finally, we sum up two loss functions with the weighting parameter $\lambda$, and the final loss function is formulated as follows:
\begin{align}
\mathcal{L} &= \mathcal{L}_{\text{tea}} + \lambda \cdot \mathcal{L}_{\text{reblur}}\\
&= \sum_{m=1}^{M}\mathcal{L}_{\text{LPIPS}}(\mathbf{L}_{\text{bsn}}^{\uparrow}(t_m),\mathbf{L}_{\text{ldn}}(t_m)) + \lambda \cdot \mathcal{L}_{\text{MSE}}(\widetilde{\mathbf{B}},\mathbf{B}) .
\end{align}

\begin{figure}[t]
    \centering
    \includegraphics[width=1\linewidth]{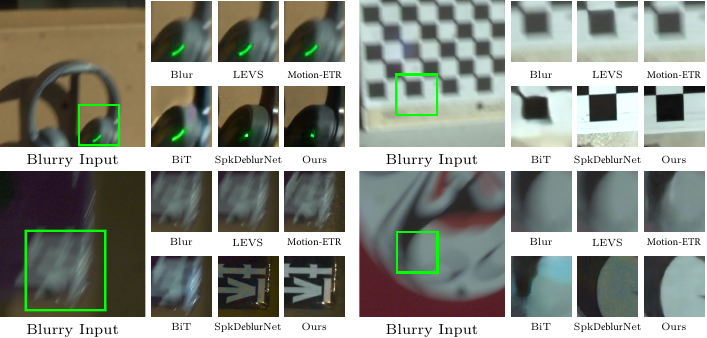}
    \vspace{-1.4em}
    \caption{Qualitative comparison for the single frame restoration on the RSB dataset.}
    \label{fig:compare_rsb1}
    \vspace{-1em}
\end{figure}
\begin{figure}[t]
    \centering
    \includegraphics[width=1\linewidth]{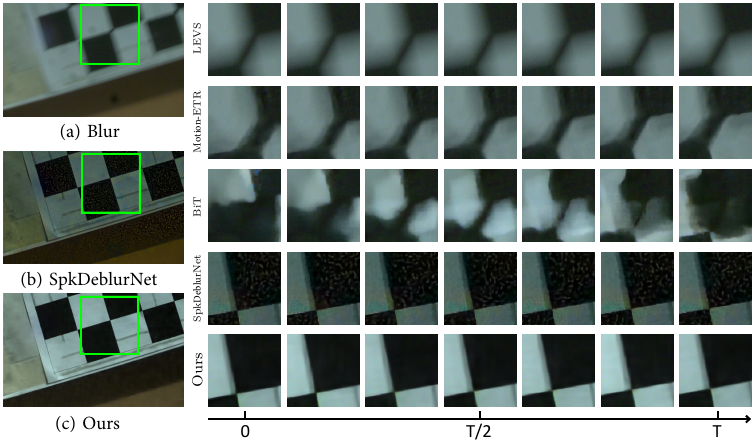}
    \vspace{-1em}
    \caption{Qualitative comparison for the sequence reconstruction on the RSB dataset. }
    \label{fig:compare_sequence}
    \vspace{-1.7em}
\end{figure}

\section{Experiments} \label{sec:experiments}
\subsection{Dataset}
\noindent\textbf{Synthetic Data.} For quantitative analysis of our spike-guided motion deblurring task, we construct the synthetic dataset based on the widely employed GOPRO \cite{GOPRO} dataset. We initially utilize the interpolation algorithm XVFI \cite{sim2021xvfi} to augment the video frame by interpolating additional $7$ frames between each pair of consecutive sharp images. To generate the spike stream that mimics reality closely, we downsample the interpolated video to the resolution of $320\times 180$ and simulate the spike stream based on the spike simulator \cite{zhao2022spikingsim}. To replicate real-world motion blur, we synthesize each blurry input by averaging $97$ frames from interpolated video sequences. 
% As for dataset division, we adhere to the official GOPRO dataset division, which includes 22 video sequences for training and 11 for testing.

\noindent\textbf{Real-world Data.} 
% To comprehensively assess the deblurring performance of our S-SDM in real-world scenarios, 
We construct an RGB-Spike binocular system and propose the first \textbf{R}eal-world \textbf{S}pike-guided \textbf{B}lur dataset (\textbf{RSB}) in this community. This system consists of a fixed-exposure RGB camera (Basler acA1920-150uc) and a spike camera \cite{spikecamera}, enabling us to capture the blurry image and the corresponding spike stream simultaneously. Further details about our RSB and the spatial-temporal calibration for our binocular system are provided in the supplementary materials.

\begin{table*}[t]
\footnotesize
\caption{Quantitative comparison of the sequence reconstruction task on the GOPRO dataset.}

\vspace{0.5em}
\centering
 \resizebox{\linewidth}{!}{%
\begin{tblr}{
 width = \textwidth,
  colspec = {X[5,l]X[1,c]X[0.05,c]X[1.3,c]X[1.3,c]X[0.05,c]X[1.3,c]X[1.3,c]X[0.05,c]X[1.3,c]X[1.3,c]X[1,c]},
  cell{1}{1} = {r=2}{},
  cell{1}{2} = {r=2}{},
  cell{1}{4} = {c=2}{},
  cell{1}{7} = {c=2}{},
  cell{1}{10} = {c=2}{},
  cell{1}{12} = {r=2}{},
  hline{3,10} = {-}{},
  hline{2} = {4-5}{},
  hline{2} = {7-8}{},
  hline{2} = {10-11}{},
  hline{1} = {1-12}{2pt},
  hline{Z} = {1-12}{2pt}, 
}
Methods      & Spike &  & $V_{th}$=1 &     &     & $V_{th}$=2 &    &      & $V_{th}$=4 &      & Params \\
            & &  &  PSNR  & SSIM  &    & PSNR & SSIM &     & PSNR & SSIM &        \\
LEVS\cite{levs}        &  $\times$ & &     21.155 & 0.601    &   &   21.155 & 0.601       &      &      21.155 & 0.601    & 4.97M       \\
Motion-ETR\cite{motion_etr}  & $\times$ & &      21.955 & 0.610   &    &      21.955 & 0.610      &      &     21.955 & 0.610       &    6.55M    \\
BiT\cite{bit}         & $\times$ & &     23.644 & 0.698  &    &        23.644 & 0.698     &      &     23.644 & 0.698      &    11.3M    \\
TRMD\cite{trmd}+DASR\cite{DASR}  & \checkmark  &  & 27.323     &  0.784 &    &    21.198   &   0.601   &      &     18.567 &  0.523    &   19.3M     \\
RED\cite{red}+DASR\cite{DASR}     &  \checkmark & &  24.456     &   0.741 &    &   23.178   &  0.674    &      &  21.942    &   0.608   &    9.76M    \\
REFID\cite{refid}+DASR\cite{DASR}     &  \checkmark & &    28.124   &  0.819  &    &  15.288    &    0.339   &      &  13.623    &    0.274   &    15.9M    \\
SpkDeblurNet\cite{spike_deblur} &  \checkmark  &  &\textbf{28.307}  & \textbf{0.834}  &   &  14.406  &  0.299     &      &  11.621    &   0.202       &    13.4M    \\
S-SDM (Ours)      &   \checkmark   &  &  26.893  & 0.757  &    &  \textbf{26.367}     &    \textbf{0.740}  &      &  \textbf{25.433}    &    \textbf{0.699}  &    \textbf{0.23M }   
\end{tblr}
}
\label{tab:GOPRO_result}
% \vspace{-1.5em}
\end{table*}        
% 放支撑材料，上面相机的内容也是
% \noindent\textbf{Implementation Details.} Our method requires separate training of the blind-spot network, super-resolution network, and deblurring network. We first completed the training of the BSN network on the low spatial resolution pulse stream of the GOPRO dataset, using a random cropping strategy of size $128\times 128$, an initial learning rate of $3e^{-4}$ with the Adam optimizer, and training for 1000 epochs on an NVIDIA GeForce GTX 3090 GPU with a batch size of 4. Then, in the blurred-pulse pair data, we trained the super-resolution network for 70 epochs with the same training settings as BSN. Finally, Deblur-Net was trained for 100 epochs with a learning rate of $1e^{-3}$ using the distillation learning framework. For quantitative analysis of our method's performance on the synthetic dataset, we used common metrics in the image deblurring field: PSNR, SSIM, and LPIPS. In the real dataset, our training procedure was consistent with the above, and we analyzed the performance of different methods from a qualitative perspective.

\begin{figure}[t]
    \centering
    \includegraphics[width=1.0\linewidth]{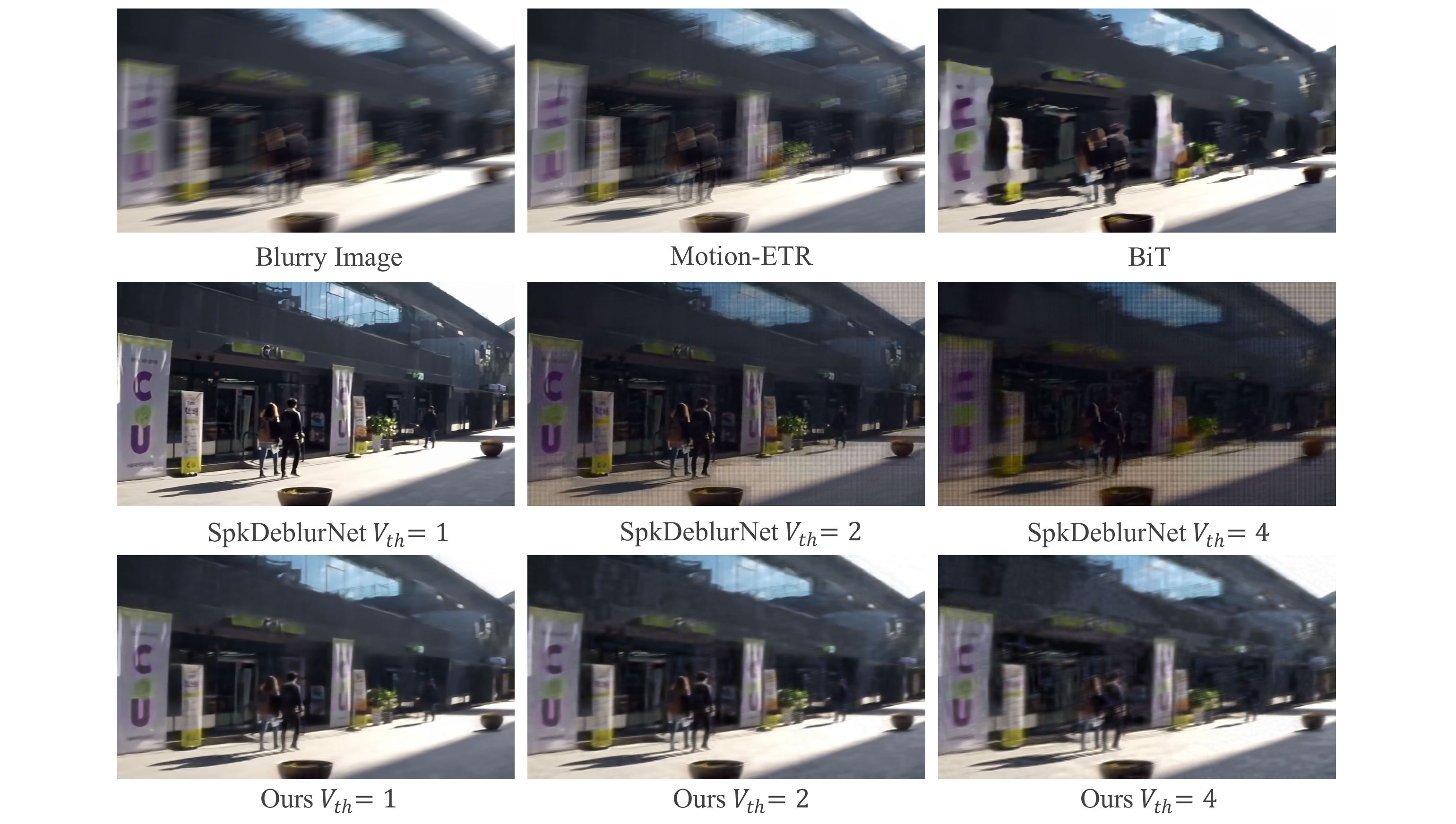}
    \vspace{-1em}
    \caption{Visual comparison of our S-SDM against other methods on the GOPRO dataset.}
    \label{fig:gopro3_compare}
    \vspace{-1em}
\end{figure}

\subsection{Experimental Results}
We conduct both quantitative and qualitative comparisons of our S-SDM against state-of-the-art (SOTA) motion deblurring methods, including frame-based Motion-ETR \cite{motion_etr}, LEVS  \cite{levs}, video-based BiT \cite{bit}, event-based TRMD \cite{trmd}, REFID \cite{refid}, RED \cite{red} and spike-based SpkDeblurNet \cite{spike_deblur} on the GOPRO and RSB datasets. 
% Since SpkDeblurNet does not fully consider the real spatial-resolution gap between two modalities in real-world scenarios, we modify its spike down-sampling module to perform convolution at the original size. 
For event-based methods, we replace the event stream with the spike stream and adopt the same input representation in these methods \cite{red,refid,trmd}. We further cascade the image super-resolution technique  DASR \cite{DASR} as in \cite{gem} for the deblurred sequence to overcome the modality resolution inconsistency which is not considered in these methods. We reconstruct $7$ images from one blurry input for sequence restoration evaluation \cite{gem} as listed in \cref{tab:GOPRO_result}. 

\textbf{Results on GOPRO.} To simulate the spike density domain gap as depicted in \cref{fig:fig1}, we train all supervised spike-based deblurring methods on the GOPRO dataset under spike threshold \cite{zhao2022spikingsim} $V_{th} = 1$ and evaluate them on datasets with spike thresholds $V_{th} = 1, 2, 4$. Quantitative comparison results are listed in \cref{tab:GOPRO_result} and the visual comparison is demonstrated in \cref{fig:gopro3_compare}. Given that the principal contribution of our S-SDM is self-supervised learning, it is foreseeable that our method might be slightly inferior to the supervised methods SpkdDeblurNet on the dataset with $V_{th} = 1$. While these supervised methods deteriorate on datasets with $V_{th} \neq 1$, our method achieves great generalization benefiting from the self-supervision design. Specifically, SpkDeblurNet tends to produce darker and blurrier reconstructions on images with high thresholds as shown in \cref{fig:gopro3_compare}. Besides, our method achieves better restoration performance than the self-supervised method RED due to our consideration of the spatial-resolution mismatch between two modalities and the designed teacher loss, which imposes a stronger constraint than the optical loss in RED.

% Revisiting the mentioned motion ambiguity problem, we observe that frame-based methods LEVS and Motion-ETR suffer significant performance degradation in transitioning from single frame restoration to sequence reconstruction, with decreases of 4.425 dB $\downarrow$ and 5.866 dB $\downarrow$ in PSNR respectively. In contrast, our method alleviates this issue benefiting from the incorporation of the high temporal-resolution spike stream.

% While other supervised methods are trained on the synthetic dataset and then evaluated on the real-world, our method is further re-trained on the RSB dataset, which shows notable distinctions from the GOPRO dataset in aspects such as spike density, spike noise, and blur generation mode. 

\textbf{Results on RSB.} We further present visualizations of single frame and sequence reconstruction comparisons on the real-world RSB dataset as depicted in \cref{fig:compare_rsb1,fig:compare_sequence}. Both frame-based and video-based approaches fail to replicate fine textures and detailed elements present in the blurry input. While SpkDeblurNet is capable of recovering structural details and the motion trajectory, it is deteriorated by significant noise and color distortion under conditions with lower spike density than those simulated in the GOPRO dataset. Similarly, in scenarios of higher spike stream densities, the restoration of SpkDeblurNet tends to exhibit over-exposure, resulting in brightness inconsistency compared to the blurry input. This over-exposure affects the dynamic range of the reconstructed image, ultimately compromising the overall perceptual quality and uniformity of the restored sequence. Our method addresses these challenges by finetuning on the RSB dataset, ensuring that the restored sequence aligns with the real-captured blurry input and spike stream. More comparative experiments and analyses are accessible in the supplementary material.  

% In terms of high-frequency details, other methods struggle to eliminate the laser ghost as evident in the earphone scene of \cref{fig:compare_rsb1}. 
% todo 补充材料
% In the qualitative experiments focusing on sequence restoration as in \cref{fig:compare_sequence}, LEVS fails to analyze motion information during the exposure, resulting in uniform outcomes across each image in the sequence. While Motion-ETR and BiT appear to trace the moving pattern of the calibration board correctly, the texture restoration is severely destroyed. Benefiting from the incorporation of the spike stream, though SpkDeblurNet correctly recovers the trajectory, it encounters deterioration problems similar to those observed in \cref{fig:compare_sequence}. Our method tackles the aforementioned challenges in a self-supervised manner, which facilitates re-training on the real-blur dataset, ensuring that the restored sequence is aligned with the real-captured blurry input and spike stream. 

\subsection{Ablation Study}
We perform ablation experiments on the GOPRO and RSB datasets to evaluate the performance of each module within S-SDM, the validity of our designed network architecture, as well as the overall effectiveness of our distillation learning framework. In this subsection, we evaluate the performance based on the single middle frame for simplicity.

\textbf{Modules Cascading.} Building upon the SDM, we sequentially cascade the BSN, the SR, and the LDN to evaluate their respective effectiveness, with quantitative results on GOPRO presented in \cref{tab:ablation1}. We employ bilinear interpolation as a substitute for the SR network in experiments I-1 and I-2 to align the spatial resolution of two modalities. 

\begin{figure}[H]
    \centering
    \begin{minipage}{0.49\textwidth}
        \centering
        \includegraphics[width=\textwidth]{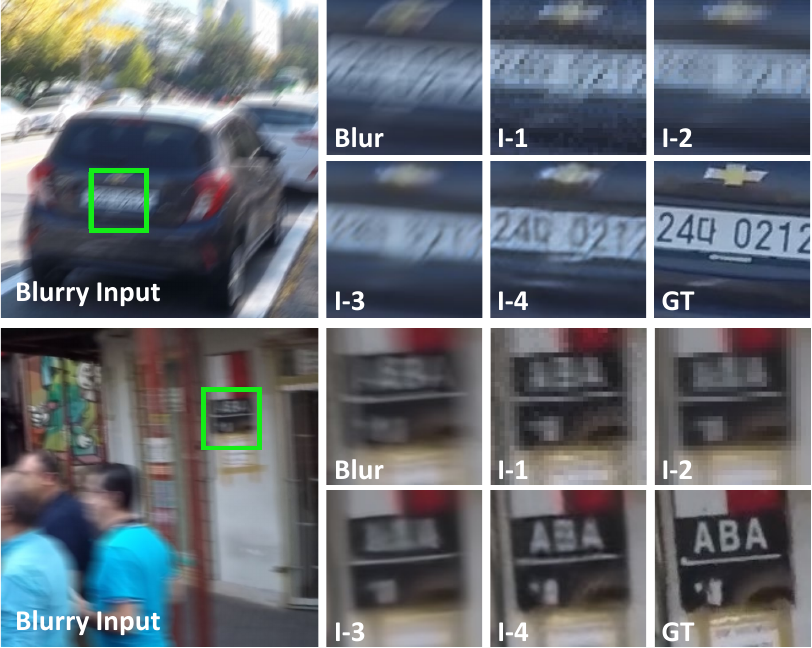}
    \vspace{-1.5em}
        \caption{Modules cascading comparisons on GOPRO. Experiments ID can be viewed on \cref{tab:ablation1}.
        % Experiments corresponding to the ID can be viewed through \cref{tab:ablation1}.
        }
        \label{fig:ablation1}
    \end{minipage}
    \hfill
    \begin{minipage}{0.485\textwidth}
        \centering
        \includegraphics[width=\textwidth]{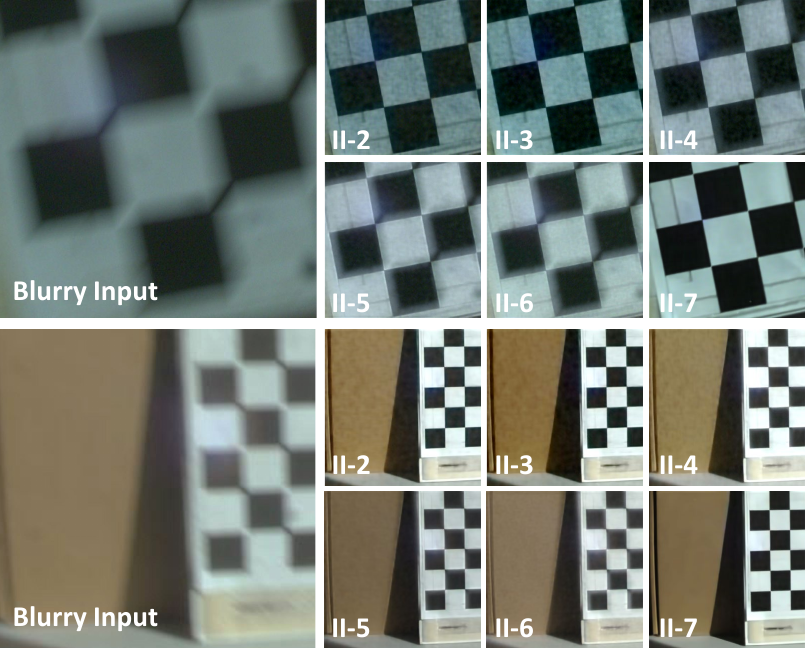}
    \vspace{-1.5em}
        \caption{Distillation learning comparisons on RSB. Experiments ID can be viewed on \cref{tab:ablation2}
        }
        \label{fig:ablation2}
    \end{minipage}
    \vspace{-1.2em}
\end{figure}

Qualitative ablation experiments are illustrated in \cref{fig:ablation1}. These comparisons reveal that while the SDM effectively removes motion blur, it struggles with significant noise and detail loss due to the spike noise and the low resolution of the spike stream. While the BSN mitigates noise and the SR network improves spatial resolution explicitly, the LDN trained via distillation learning further refines these enhancements, enabling the recognition of intricate textural features in the images, such as the license plate and the door number shown in \cref{fig:ablation1}.

\begin{table}[t]
    \centering
    \caption{Performance comparison between the SAN in GEM \cite{gem} and our designed LDN.}
    \label{tab:gem_ldn}
    \begin{tabular}{lcccc}
        \toprule[2pt]
        Methods & PSNR $\uparrow$ & SSIM $\uparrow$ & Params (M) $\downarrow$ & Flops (G) $\downarrow$\\
        \midrule
        SAN \cite{gem} & 27.283 & 0.773 & 2.36 & 107.84 \\
        LDN (Ours) & \textbf{27.928} & \textbf{0.786} & \textbf{0.234} & \textbf{33.60} \\
       \bottomrule[2pt]
    \end{tabular}
    % \vspace{-1em}
\end{table}

\textbf{Network Architecture.} 
While our designed LDN mirrors the Scale-aware Network (SAN) proposed in the GEM \cite{gem}, we replace the SAN with the LDN to compare the performance difference between the two architectures as depicted in \cref{tab:gem_ldn}. In the self-supervised learning framework discussed in this paper, enhancing restoration performance primarily hinges on improving the quality of pseudo-labels rather than the network architecture itself. Despite its simple design, which consists only of convolutional layers, ResBlocks, and basic modules such as CBAM \cite{CBAM}, LDN outperforms SAN in both PSNR and SSIM while requiring fewer parameters and less computation, demonstrating that LDN is both sufficient and efficient for the self-supervised spike-guided motion deblurring task.

\begin{table}[H]
\centering
\setlength{\tabcolsep}{4pt} % Adjust column separation
\begin{minipage}[t][0.25\linewidth][c]{0.49\linewidth}
  \centering
  \caption{Modules cascading ablation on GOPRO.}
  \label{tab:ablation1}
\vspace{-0.5em}
  \resizebox{\linewidth}{!}{%
  \large % Increase font size
  \begin{tabular}{
    >{\centering\arraybackslash}p{0.15\linewidth}
    *{3}{>{\centering\arraybackslash}p{0.17\linewidth}}
    *{2}{>{\centering\arraybackslash}p{0.17\linewidth}}
  }
    \toprule[2pt]
    ID & BSN & SR & LDN  & PSNR$\uparrow$ & SSIM$\uparrow$ \\
    \midrule
    I-1 & $\times$ & & & 23.012 & 0.486 \\
    I-2 & $\checkmark$ & & & 24.634 & 0.661 \\
    I-3 & $\checkmark$ & $\checkmark$ & & 26.144 & 0.708 \\
    I-4 & $\checkmark$ & $\checkmark$ &  $\checkmark$ & \textbf{27.928} & \textbf{0.786} \\
    \bottomrule[2pt]
  \end{tabular}%
  }
\end{minipage}\hfill
\begin{minipage}[t][0.25\linewidth][c]{0.49\linewidth}
  \centering
  \caption{Distillation learning ablation on GOPRO.}
  \label{tab:ablation2}
\vspace{-0.5em}
  \resizebox{\linewidth}{!}{%
  \large % Increase font size
  \begin{tabular}{
    >{\centering\arraybackslash}p{0.15\linewidth}
    *{3}{>{\centering\arraybackslash}p{0.17\linewidth}}
    *{2}{>{\centering\arraybackslash}p{0.17\linewidth}}
  }
    \toprule[2pt]
    ID & $\mathcal{L}_{\text{tea}}$ & $\mathcal{L}_{\text{reblur}}$ & $\lambda$ & PSNR$\uparrow$ & SSIM$\uparrow$ \\
    \midrule
    II-1  & $\times$       & $\checkmark$     & 10 & 23.102 & 0.441 \\
    II-2  & $\checkmark$   & $\times$ & / & 26.563 & 0.723 \\
    II-3  & $\checkmark$   & $\checkmark$     & 10 & 27.345 & 0.762 \\
    II-4  & $\checkmark$   & $\checkmark$     & 50 & 27.742 & 0.778 \\
    II-5  & $\checkmark$   & $\checkmark$     & 100 & \textbf{27.928} & \textbf{0.786} \\
    % II-5  & $\checkmark$   & $\checkmark$     & 150 & 27.841 & 0.785 
    II-6  & $\checkmark$   & $\checkmark$    & 200 & 27.620 & 0.783 \\
    % II-6  & $\checkmark$   & $\checkmark$    & 250 & 27.620 & 0.783 \\
    \bottomrule[2pt]
  \end{tabular}%
  }
\end{minipage}
\vspace{-1.3em}
\end{table}

\textbf{Distillation Learning.} We focus on analyzing the contribution of the teacher loss $\mathcal{L}_{\text{tea}}$ and the reblur loss $\mathcal{L}_{\text{reblur}}$ within the distillation framework, with quantitative results listed in \cref{tab:ablation2}. Without the teacher loss $\mathcal{L}_{\text{tea}}$, the LDN tends toward learning identity mapping from the blurry input. While under the guidance of the teacher model, the reblur loss $\mathcal{L}_{\text{reblur}}$ not only enforces motion consistency in the reconstructed sequence but also enriches the LDN with high-resolution details from the non-blurry regions of the input, thus improving the performance on GOPRO as listed in \cref{tab:ablation2}.
% Different from the reconstruction loss designed in EVDI \cite{evdi}, which utilizes two adjacent blurry images for mutual constraint, our reblur loss relies solely on a single blurry frame. Therefore, it becomes evident that the LDN tends toward learning identity mapping from the blurry input without supervision from the teacher model. Nevertheless, under the guidance of the teacher model, the reblur loss $\mathcal{L}_{reblur}$ plays a dual role. It not only enforces motion consistency in the reconstructed sequence but also enriches the LDN with high-resolution details from the non-blurry regions of the blurry input. Quantitative evaluations conducted on the GOPRO as detailed in \cref{tab:ablation2} demonstrate that the incorporation of reconstruction loss enhances the performance of LDN. Notably, within a specific range, an increase in the hyperparameter $\lambda$ correlates with improved performance.

We further apply the LDN trained on GOPRO to the real-world dataset RSB, with qualitative visualization illustrated in \cref{fig:ablation2}. As observed in the figure, the absence of the reblur loss $\mathcal{L}_{\text{reblur}}$ leads to significant noise in the recovered image, which predominantly arises from the disparity in spike density and generation mechanism between the simulated and real-captured spike stream. This discrepancy causes the LDN to overestimate the spike number, resulting in black holes in regions with lower spike density than simulated, which reflects the drawback inherent in the supervised learning strategies as discussed in \cref{sec:intro}. Increasing the weight of the reblur loss $\mathcal{L}_{\text{reblur}}$ allows the LDN to incorporate more information from the blurry input, thereby mitigating this issue. However, this adjustment also leads to the presence of blurry edges. We follow the parameters set in Experiment II-6 and retrain the LDN on the RSB dataset (referred to as Experiment II-7). The retrained LDN effectively recovers the sharp edge of the calibration board and suppresses the spike noise in the background, validating the feasibility of our self-supervised framework in real-world scenarios.

\section{Conclusion}
In this paper, we introduce a novel self-supervised spike-guided motion deblurring framework S-SDM, which reconstructs sequences of sharp images from real-world blurry inputs with the spike stream. Additionally, we construct an RGB-Spike binocular system and propose the first spatially-temporally calibrated real-world dataset RSB in this community. Quantitative and qualitative experiments validate the superior generalization capabilities of our proposed S-SDM.

\textbf{Limitation.} The limitation of our S-SDM lies in its dependence on strict spatial-temporal calibration. Misalignment will lead to color shifts and quality degradation in the deblurred sequence. 
% In the future, we plan to develop a self-supervised deblurring method that does not require strict spatial-temporal calibration, thereby enhancing its applicability in real-world scenarios.

\clearpage  % TODO REVIEW/FINAL: This \clearpage needs to be removed from both review and camera-ready versions.

% \bibliographystyle{plainnat}
% {\small
% \bibliographystyle{ieee_fullname}
% \bibliography{main}
% }
\newpage
\section*{Acknowledgments}
We sincerely appreciate Yuyan Chen (HUST) for her valuable suggestions and for polishing the figures. This work was supported by the National Natural Science Foundation of China (62176003, 62088102, 62306015), the China Postdoctoral Science Foundation (2023T160015), the Young Elite Scientists Sponsorship Program by CAST (2023QNRC001), and the Beijing Nova Program (20230484362). 

{\small
\bibliographystyle{ieee_fullname}
\bibliography{main}
}

\newpage
\appendix
\section{Appendix}

\begin{figure}[t]
    \centering
    % First animation
    \begin{minipage}{.99\linewidth}
        \centering
        \animategraphics[width=\linewidth, autoplay, loop]{3}{imgs/Sup/middle/middle_}{0}{29}
    \end{minipage}
    \\
    \vspace{1em}
    % Second animation
    \begin{minipage}{.99\linewidth}
        \centering
        \animategraphics[width=\linewidth, autoplay, loop]{3}{imgs/Sup/high/high_}{0}{29}
    \end{minipage}

    % Shared caption and label for both animations
    \caption{Video comparison of our method against the BiT and SpkDeblurNet on the RSB dataset under different luminance conditions, with the static visualization displayed in \cref{sup_fig:compare_sequence_together}. It is recommended to view the pdf using Acrobat PDF reader and the gif demonstration with a higher frame rate is available in the supplementary zip file.}
    \label{sup_fig:animations}
    \vspace{-1em}
\end{figure}

This supplementary material provides a comprehensive elaboration on the methodologies and experiments in this paper. It is organized into four distinct sections: Theory Analysis in \cref{sec_sup:theory}, Network Settings in \cref{sec_sup:network}, RSB Dataset in \cref{sec_sup:dataset}, Experimental Details in \cref{sec_sup:experiments} and Additional Figure in \cref{sec_sup:figures}.

\subsection{Theory Analysis} \label{sec_sup:theory}
% \subsubsection{Spike Camera}
% Consider $L(t)$ to represent the latent sharp frame at time $t$. Each pixel $p$ in the spike camera has an integrator that accumulates the incoming photons at a high frequency as shown in \cref{fig_sup:spike_camera}. Once the cumulative intensity exceeds a predefined threshold $C$ at time $t_e$, pixel $p$ emits a spike, and the integrator is reset to zero, formulated as:
% \begin{equation}
% \int_{t_s}^{t_e}L(t)dt \geq C, \label{equ:spike_generation}
% \end{equation}
% where $t_s$ denotes the firing time of the previous spike. Targeting objects in rapid motion, the spike camera generates the spike bit stream with extremely low latency as demonstrated in \cref{fig_sup:spike_frame}.
% \subsubsection{Spike-guided Deblur Model}
 We define $k_1^{\mathbf{B}}$ and $k_2^{\mathbf{B}}$ as the ratios of the red channel to the green and blue channels in the blurry input $\mathbf{B}$ respectively, \ie,
\begin{align}
    k_1^{\mathbf{B}} &= \mathbf{B}_r / \mathbf{B}_{gre}, \\ 
    k_2^{\mathbf{B}} &= \mathbf{B}_r / \mathbf{B}_{b}.
\end{align}
Since the gray image is the weighted sum of RGB channels, the ratio of the gray to the red image $\alpha_r^{\mathbf{B}}$ is formulated as:
\begin{align}
\alpha_r^{\mathbf{B}} &= \mathbf{B}_g / \mathbf{B}_r \\
&= (w_r\cdot \mathbf{B}_r+w_{gre}\cdot \mathbf{B}_{gre}+w_b\cdot \mathbf{B}_b) / \mathbf{B}_r \\
&= w_r + w_{gre} / k_1^{\mathbf{B}} + w_{b} / k_2^{\mathbf{B}} \label{equ_sup:1} ,
\end{align}
where $w_c$ denote the weight of channel $c \in\{r,gre,b\}$ respectively.

Similarly, we define $k_1^{\mathbf{E}}(t,\mathcal{T}')$ and $k_2^{\mathbf{E}}(t,\mathcal{T}')$ as the fractions of the red channel relative to the green and blue channels in the short-exposure image $\mathbf{E}(t,\mathcal{T}')$, resulting in:
\begin{equation}
\alpha_r^{\mathbf{E}}(t,\mathcal{T}') = w_r + w_{gre} / k_1^{\mathbf{E}}(t,\mathcal{T}')+ w_{b} / k_2^{\mathbf{E}}(t,\mathcal{T}').  \label{equ_sup:2}
\end{equation}

Given the observation that the color information of adjacent pixels often exhibits similarity under the premise of minor motion amplitude, we postulate that the colors of the blurry input $\mathbf{B}$ and the short-exposure image $\mathbf{E}(t,\mathcal{T}')$ are identical. This assumption implies that the intensity proportion among RGB channels in two images is approximately equivalent, \ie, $k_1^{\mathbf{B}} \approx k_1^{\mathbf{E}}(t,\mathcal{T}')$ and $k_2^{\mathbf{B}} \approx k_2^{\mathbf{E}}(t,\mathcal{T}')$. Drawing from \cref{equ_sup:1} and \cref{equ_sup:2}, We further deduce the following relation: 
\begin{equation}
\alpha_r^{\mathbf{B}} \approx \alpha_r^{\mathbf{E}}(t,\mathcal{T}'),
\end{equation}
which can be readily generalized across channels, leading to $\alpha_c^{\mathbf{B}} \approx \alpha_c^{\mathbf{E}}(t,\mathcal{T}')$.

\begin{figure}[t]
    \centering
    \includegraphics[width=0.7\linewidth]{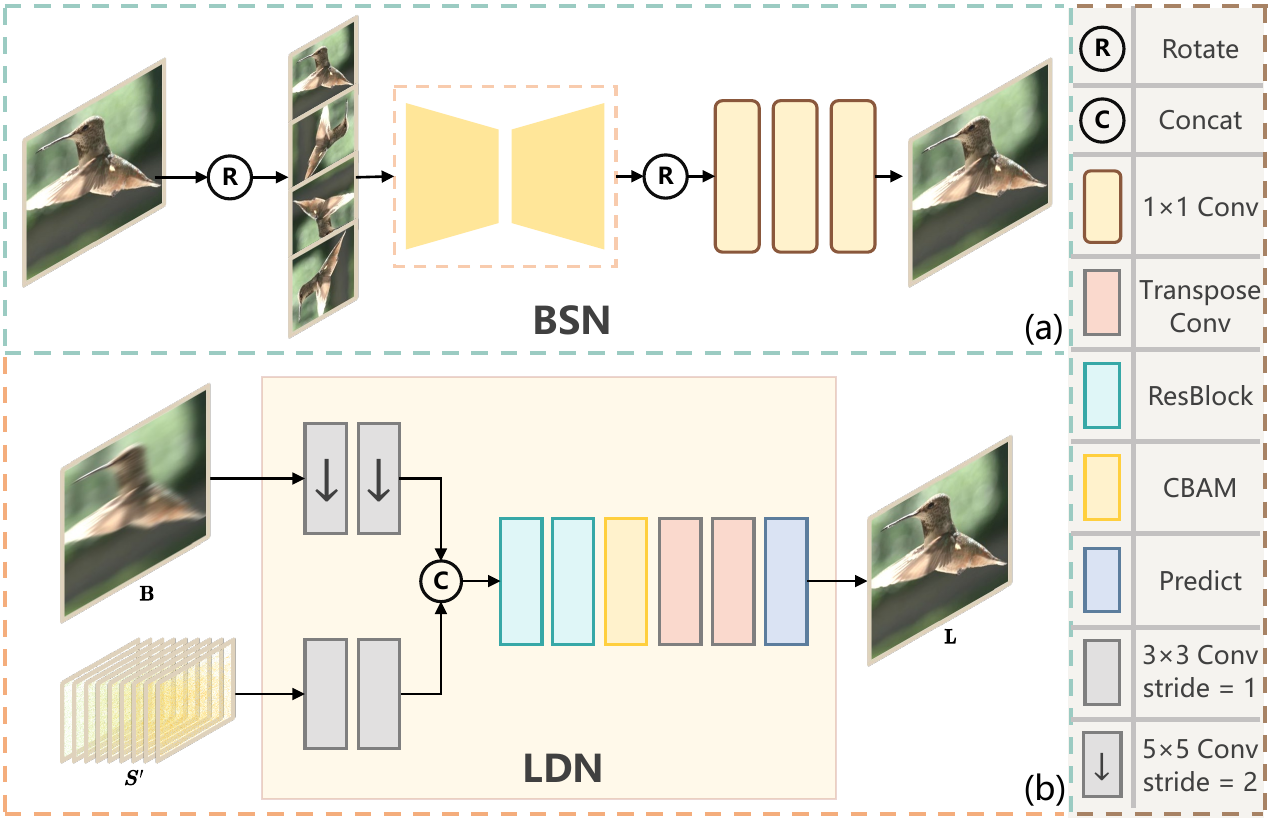}
    \caption{Network diagrams of the BSN (a) and our designed LDN (b).}
    \label{fig_sup:BSN_LDI}
    % \vspace{-1em}
\end{figure}

\subsection{Network Settings} \label{sec_sup:network}
\subsubsection{Blind Spot Network}
We construct our BSN based on the blind-spot strategy outlined in  \cite{BSN}. We first rotate the input image four times, then concatenate them into the U-Net \cite{UNet} structure like denoising network. To prevent direct mapping from the input pixel to the output pixel, a single-pixel offset strategy is employed in the convolutional kernel to separate its receptive field from the central pixel. Finally, the denoising output is obtained by merging the results of four branches via a $1\times1$ convolution, as shown in \cref{fig_sup:BSN_LDI}(a).

\subsubsection{Lightweight Deblur Network}
The network structure is depicted in \cref{fig_sup:BSN_LDI}(b), where the encoder for blurry images consists of two layers of down-sampling convolutions to align the spatial resolution of two modalities. In contrast to the intricate cross-attention mechanism outlined in  \cite{trmd,spike_deblur}, we implement the fusion of two modalities by the simple operation of $\text{Concat}(\cdot)$, with the cascaded CBAM \cite{CBAM} and residual blocks for further feature fusion. 

\subsection{RSB Dataset} \label{sec_sup:dataset}
\begin{figure}[h]
    \centering
    \begin{minipage}{0.4\textwidth}
        \centering
        \includegraphics[width=\linewidth]{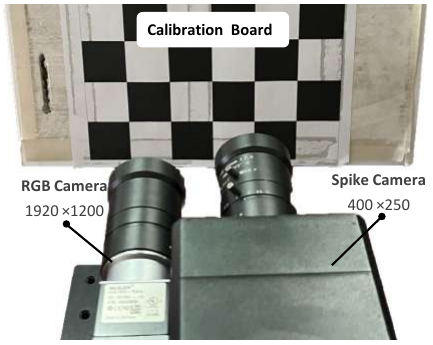}
        \caption{RGB-Spike camera system.}
        \label{fig_sup:rgb_spike_camera}
    \end{minipage}
    \begin{minipage}{0.42\textwidth}
        \centering
        \includegraphics[width=\linewidth]{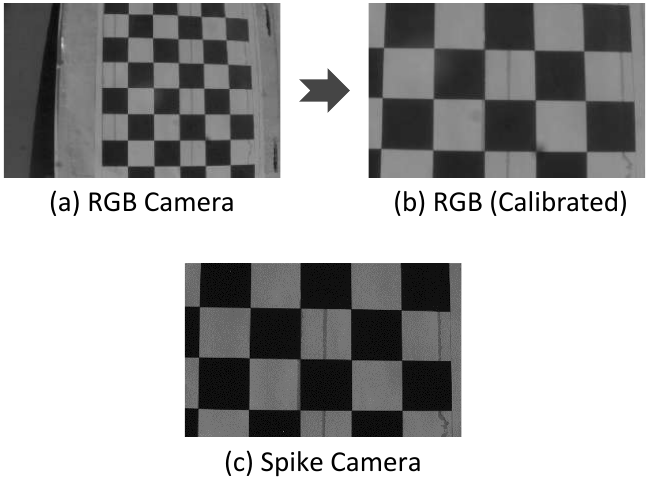}
        \vspace{-1.4em}
        \caption{Calibration result.}
        % \label{fig_sup:rgb_spike_camera}
        \label{fig_sup:calibration}
    \end{minipage}
    \vspace{-1em}
\end{figure}

We detail the construction of our RGB-Spike hybrid system as illustrated in \cref{fig_sup:rgb_spike_camera}. The system comprises a Spike Camera-001T-Gen2 with a resolution of 400 $\times$ 250 pixels, paired with a Basler acA1920-150uc RGB Camera, offering a higher resolution of 1920 $\times$ 1200 pixels.

% spike figure

 To achieve the spatial calibration of two cameras, we perform simultaneous captures of the calibration board using the RGB-Spike system as shown in \cref{fig_sup:rgb_spike_camera}. The RGB images are cropped to a resolution of 1600 $\times$ 1000, aligning them to be fourfold the resolution of the spike camera. We convert the cropped RGB image to grayscale and take the TFP \cite{tfp_tfi} image reconstruction result as the reference from the spike camera. We utilize the MATLAB  calibration toolbox to implement the calibration process, with the results detailed in \cref{fig_sup:calibration}.

\begin{table*}[t]
\centering
\caption{Quantitative comparison of the single frame task on the RSB dataset, where non-reference metric LIQE ranging from 1 to 5 is employed. LIQE is a positive metric denoted as $\uparrow$ where higher scores reflect better performance.}
% \vspace{0.5em}
\label{tab_sup:single frame}
\small % This command sets a smaller font size for the table
 \resizebox{\linewidth}{!}{%
\begin{tabular}{>{\centering\arraybackslash}p{2.5cm}>{\centering\arraybackslash}p{1.7cm}>{\centering\arraybackslash}p{1.7cm}>{\centering\arraybackslash}p{1.7cm}>{\centering\arraybackslash}p{1.4cm}>{\centering\arraybackslash}p{1.4cm}>{\centering\arraybackslash}p{1.4cm}} 
\toprule[2pt]
Methods & Board-L & Board-M & Board-H & Face & Earphone & Average $\uparrow$ \\ 
\midrule
LEVS \cite{levs}        & 1.0010    & 1.0342    & 1.0400     & 1.0029 & 1.1382 & 1.0433  \\
Motion-ETR \cite{motion_etr} & 1.0153    & 1.0064    & 1.0279     & 1.0023 & 1.1724 & 1.0449  \\
BiT \cite{bit}         & 1.0002    & 1.0069    & 1.0473     & 1.0138 & 1.6155 & 1.1367  \\
SpkDeblurNet \cite{spike_deblur} & \underline{1.8493}    & \underline{1.3232}    & \underline{2.2698}     & \underline{1.2210} & \underline{2.6606} & \underline{1.8648}  \\
S-SDM (Ours)& \textbf{2.0955} & \textbf{1.3872} & \textbf{2.4208} & \textbf{1.3814} & \textbf{2.7087} & \textbf{2.1987} \\
\bottomrule[2pt]
\end{tabular}
}
% \vspace{-1em}
\end{table*}

\begin{figure}[t]
    \centering
    \includegraphics[width=1.0\linewidth]{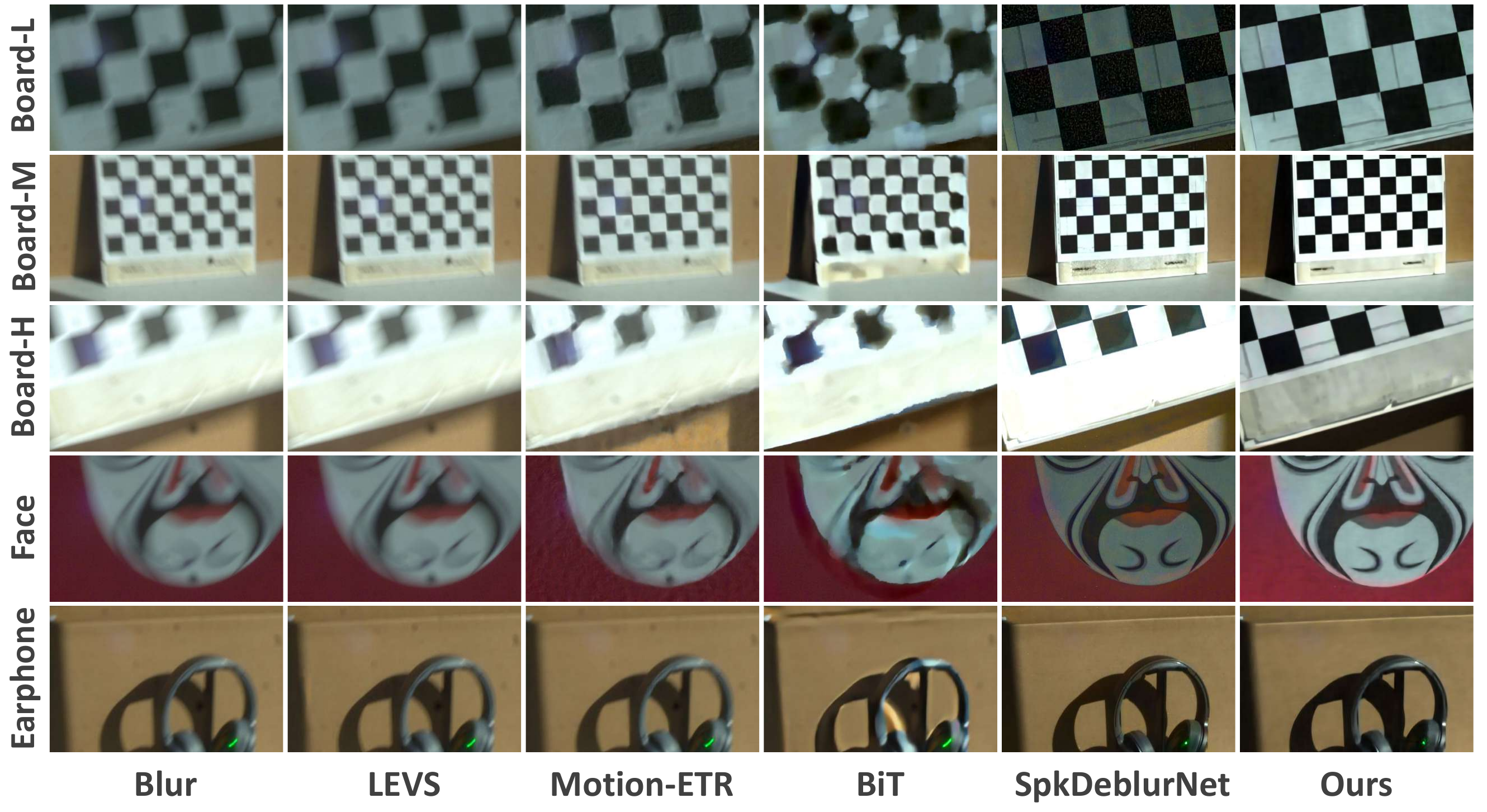}
    \caption{
    Qualitative comparisons for single-frame restoration on the RSB dataset are illustrated, where ``Board-L'', ``Board-M'', and ``Board-H'' represent the board captured under low, middle and high lighting conditions. }
    \label{fig_sup:Single_frame}
\end{figure}

Our RSB dataset contains 10 video sequences under different conditions, captured under varied conditions including scene brightness levels (\eg, Low, Middle, and High light) and motion patterns (\eg, camera shake and object motion), which introduce different types of motion blur. Besides, the RSB dataset comprises a large amount of blur-spike pairs with each blurry input corresponding to 400 spike frames.

\subsection{Experimental Details} \label{sec_sup:experiments}
\subsubsection{Comparison}
To assess the performance of our method on the GOPRO dataset, we utilize the Peak Signal-to-Noise Ratio (PSNR) and the Structural Similarity Index (SSIM) as the quantitative metrics, which are commonly used in motion deblurring tasks. In real-world datasets, where obtaining ground truth sharp sequence is challenging, we opt for the non-reference image quality assessment method Language-Image Quality Evaluator (LIQE) \cite{liqe} as a reference. By assessing visual quality through the computation of joint probabilities from visual-textual embeddings, LIQE adeptly identifies the clarity and blurriness of images independently of the ground truth, making it ideally suited for our task.

\textbf{Comparison on the RSB dataset.} Qualitative and quantitative experiments of the single frame restoration task on the RSB dataset are shown in \cref{fig_sup:Single_frame,tab_sup:single frame} respectively. The visual results coupled with the LIQE metrics demonstrate that our method outperforms other methods in handling the real-world RSB dataset. While the supervised SpkDeblurNet encounters substantial noise and overexposure challenges in both low-light and high-light environments, our approach demonstrates superior restoration performance, which is attributed to the designed self-supervised framework.

\def\imwidth{0.187}
\def\cimwid{0.125}
\def\rswidth{0.25}
\def\zuoxia{(-1.2,-0.9)}
\def\youshang{(-0.4,-0.45)}
\def\seqwidth{0.78}
\def\ssyy{(-0.8,-0.85)}
\def\ssizz{1.8cm}
\def\sswidth{0.245\textwidth}
\def\ssmag{2.5}
\def\scc{(2.12,1.4)}
\def\smallboxloc{(-0.52,-0.7)} % 绿色小方框位置
\def\largeboxloc{(2.2, 0.75)} % 绿色大方框位置

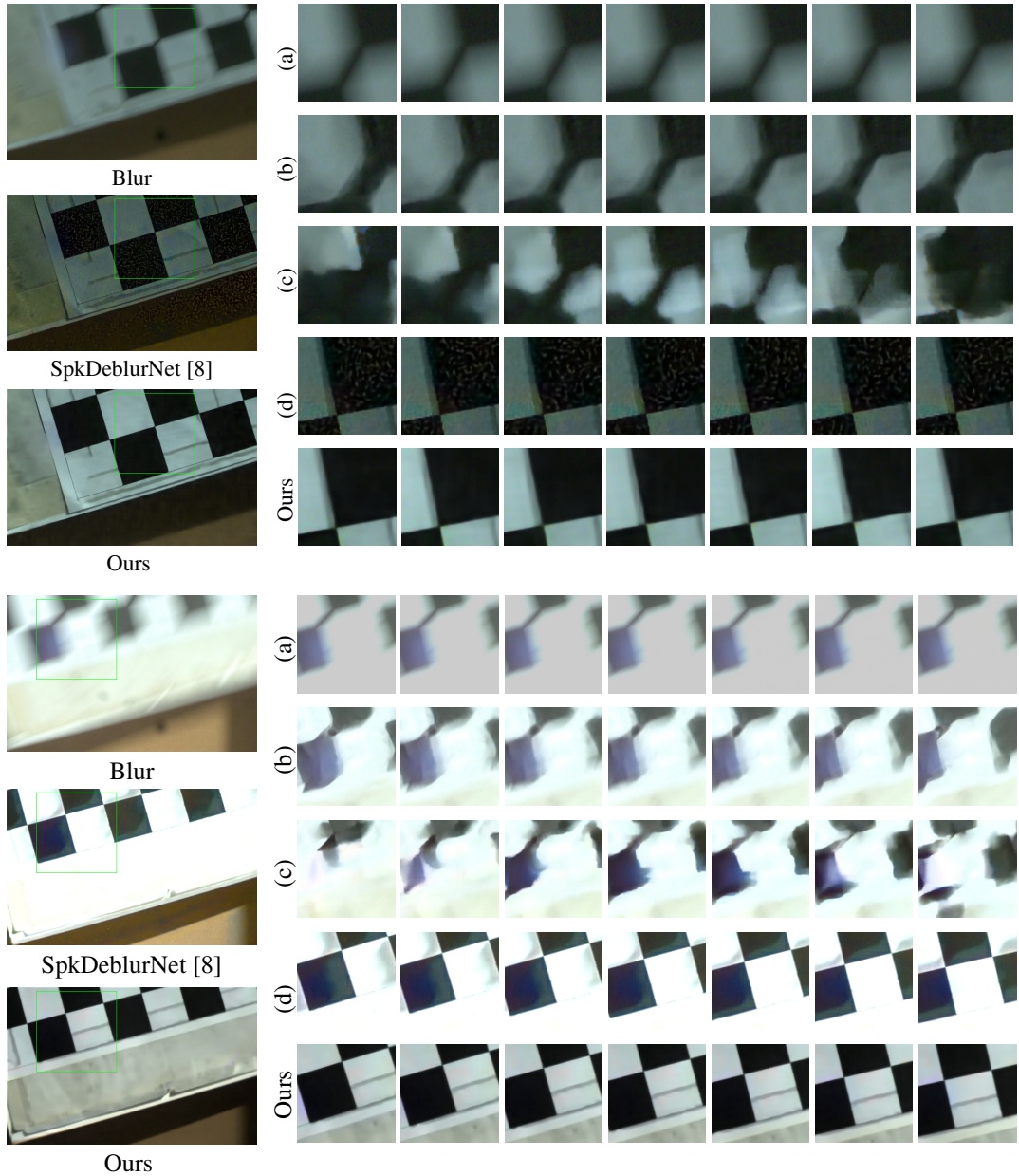
\begin{figure*}[!t]
\begin{minipage}{\textwidth}
\footnotesize
	\centering
        \begin{minipage}[t][3.9cm][t]{\rswidth\textwidth}
        \vspace{0pt}
        \centering
              \begin{tikzpicture}[spy using outlines={rectangle,green,magnification=\ssmag,size=\ssizz},inner sep=0]
        				\node {\includegraphics[width=\linewidth]{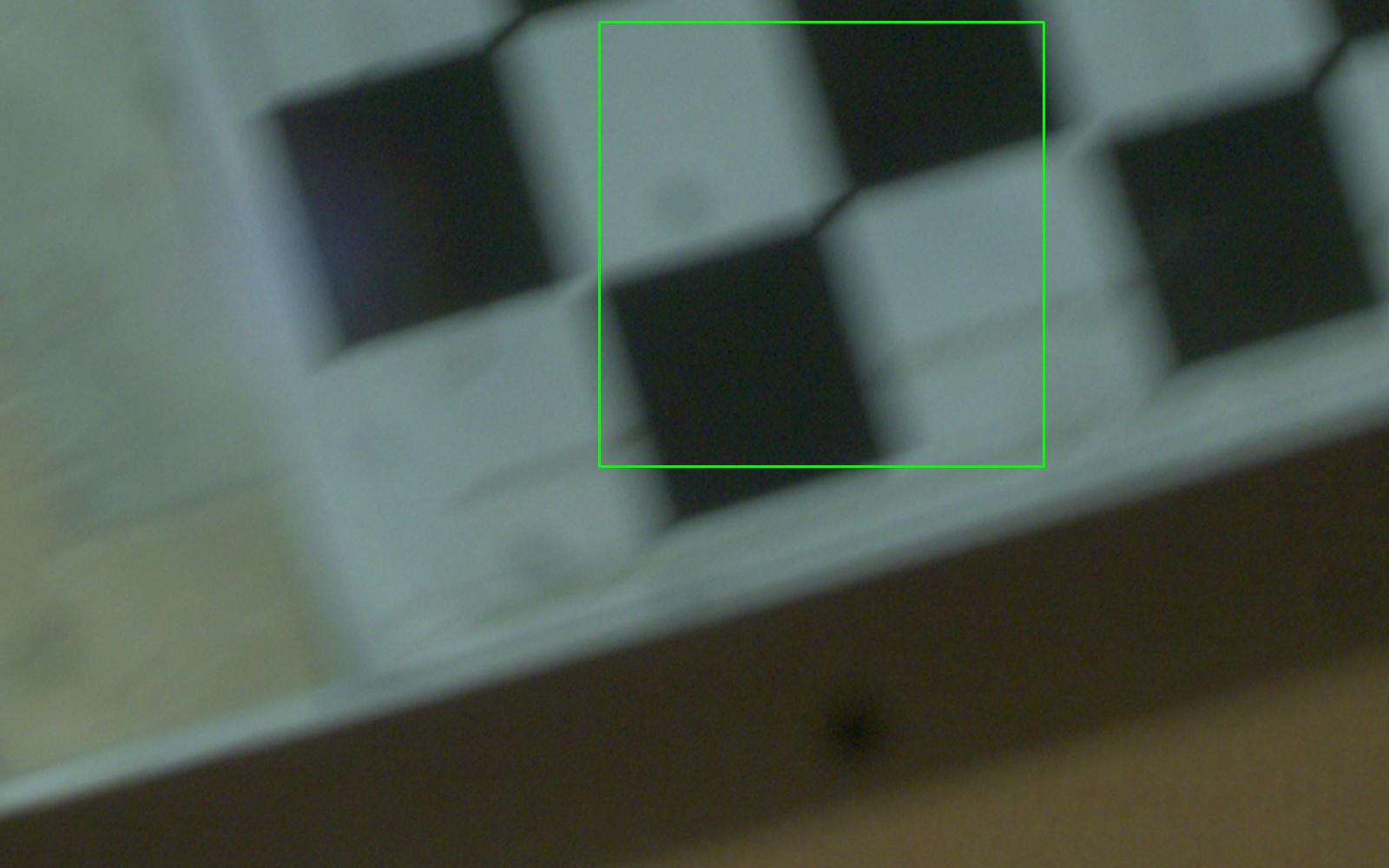}};
        				% \spy on \smallboxloc in node [left] at \largeboxloc;
        				\end{tikzpicture}
            
            Blur\vspace{0.3em}
            % \vspace{0.8em}
                    % RS frame \vspace{-0.3em}
             \vfill
			\begin{tikzpicture}[spy using outlines={rectangle,green,magnification=\ssmag,size=\ssizz},inner sep=0]
				\node {\includegraphics[width=\linewidth]{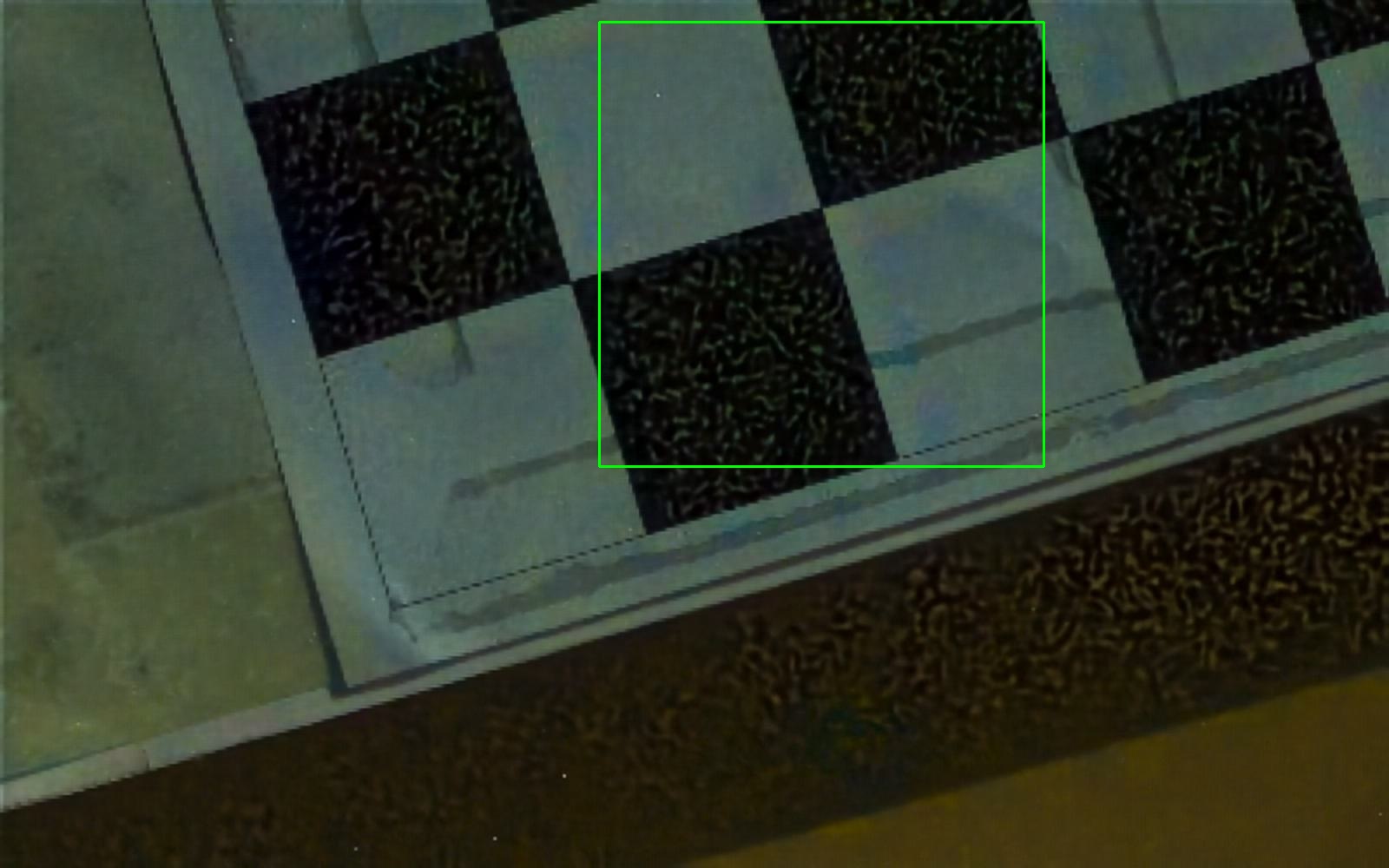}};
				% \spy on \smallboxloc in node [left] at \largeboxloc;
				\end{tikzpicture}
            
           SpkDeblurNet \cite{spike_deblur}\vspace{0.3em}
           \begin{tikzpicture}[spy using outlines={rectangle,green,magnification=\ssmag,size=\ssizz},inner sep=0]
        				\node {\includegraphics[width=\linewidth]{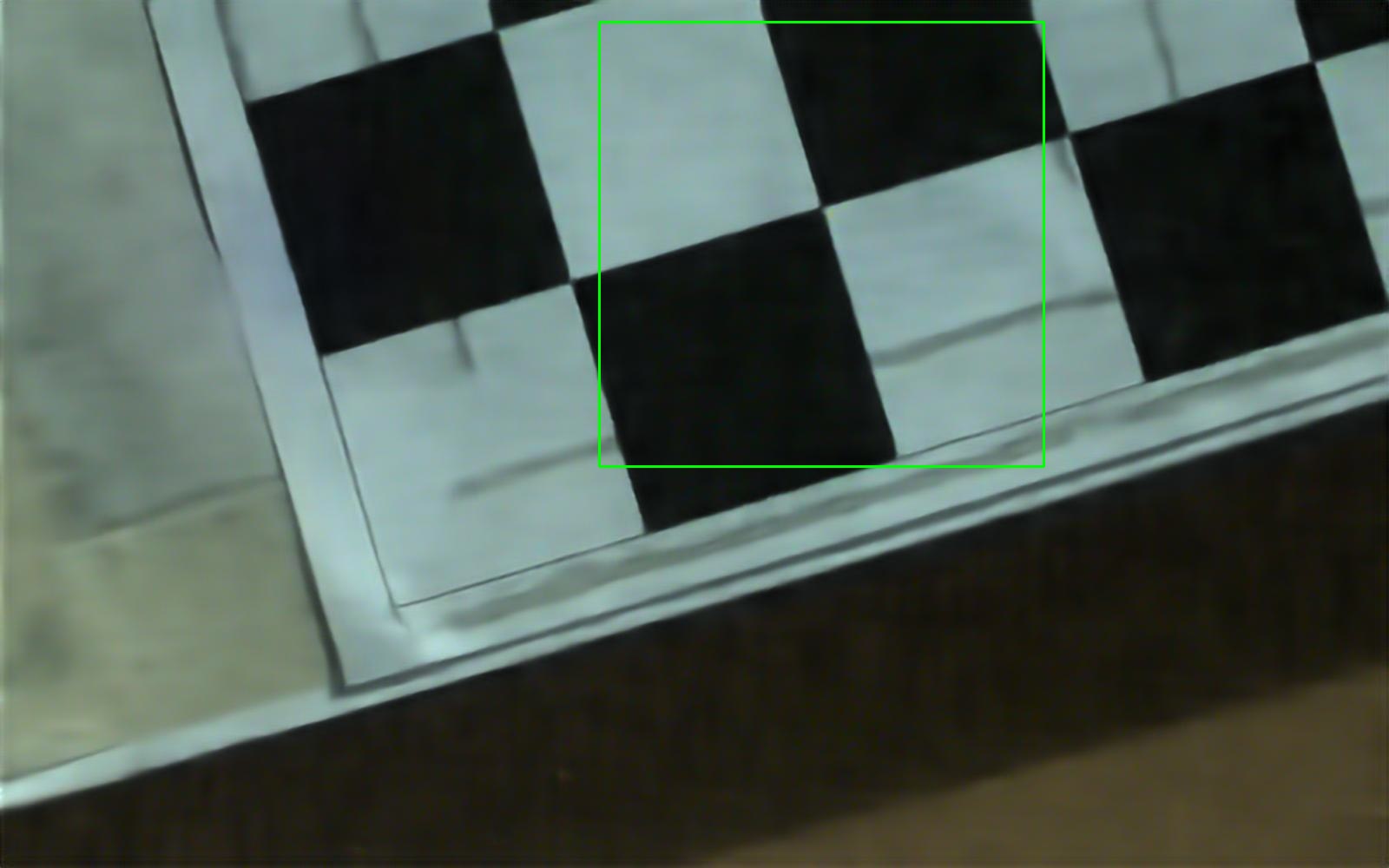}};
        				% \spy on \smallboxloc in node [left] at \largeboxloc;
        				\end{tikzpicture}
            Ours \vspace{0.3em}
    	\end{minipage}% 
% \hspace*{-1mm}
% \hspace*{-5mm}
\begin{minipage}[t]{\seqwidth\textwidth}
            \vspace{0pt}
    		\centering
      % trim={左，下，右，上}
      \begin{tikzpicture}[inner sep=0]
            \node [label={[label distance=0.4cm,text depth=0ex,rotate=90]right: \textcolor{black}{ {(a)} }}] at (0,5) {};
            \end{tikzpicture}
			\includegraphics[width=\cimwid\linewidth]{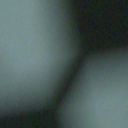}
			\includegraphics[width=\cimwid\linewidth]{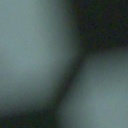}
			\includegraphics[width=\cimwid\linewidth]{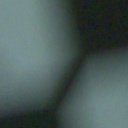}
			\includegraphics[width=\cimwid\linewidth]{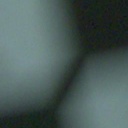}
			\includegraphics[width=\cimwid\linewidth]{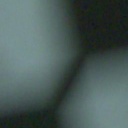}
			\includegraphics[width=\cimwid\linewidth]{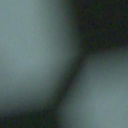}
			\includegraphics[width=\cimwid\linewidth]{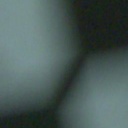}
   \\
     % \\
   \vspace{0.5em}
            \begin{tikzpicture}[inner sep=0]
            \node [label={[label distance=0.4cm,text depth=0,rotate=90]right: \textcolor{black}{ {(b)} }}] at (0,5) {};
            \end{tikzpicture}
			\includegraphics[width=\cimwid\linewidth]{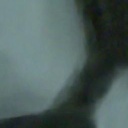}
			\includegraphics[width=\cimwid\linewidth]{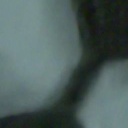}
			\includegraphics[width=\cimwid\linewidth]{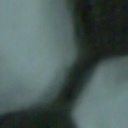}
			\includegraphics[width=\cimwid\linewidth]{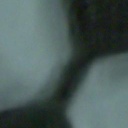}
			\includegraphics[width=\cimwid\linewidth]{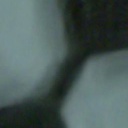}
			\includegraphics[width=\cimwid\linewidth]{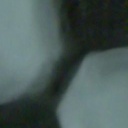}
			\includegraphics[width=\cimwid\linewidth]{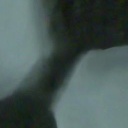}
   \\
   \vspace{0.5em}
            \begin{tikzpicture}[inner sep=0]
            \node [label={[label distance=0.4cm,text depth=-1ex,rotate=90]right: \textcolor{black}{ {(c)}}}] at (0,8.7) {};
            \end{tikzpicture}
			\includegraphics[width=\cimwid\linewidth]{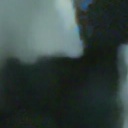}
			\includegraphics[width=\cimwid\linewidth]{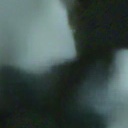}
			\includegraphics[width=\cimwid\linewidth]{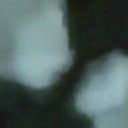}
			\includegraphics[width=\cimwid\linewidth]{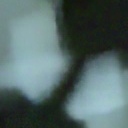}
			\includegraphics[width=\cimwid\linewidth]{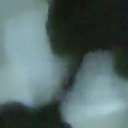}
			\includegraphics[width=\cimwid\linewidth]{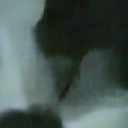}
			\includegraphics[width=\cimwid\linewidth]{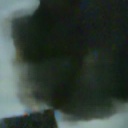}
    \\
   \vspace{0.5em}
            \begin{tikzpicture}[inner sep=0]
            \node [label={[label distance=.2cm,text depth=-1ex,rotate=90]right: \textcolor{black}{ {(d)} }}] at (-20,1) {};
            \end{tikzpicture}
			\includegraphics[width=\cimwid\linewidth]{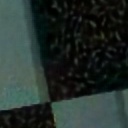}
			\includegraphics[width=\cimwid\linewidth]{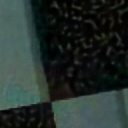}
			\includegraphics[width=\cimwid\linewidth]{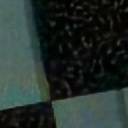}
			\includegraphics[width=\cimwid\linewidth]{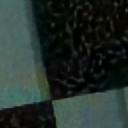}
			\includegraphics[width=\cimwid\linewidth]{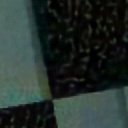}
			\includegraphics[width=\cimwid\linewidth]{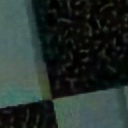}
			\includegraphics[width=\cimwid\linewidth]{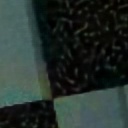}
   \\
   \vspace{0.5em}
            \begin{tikzpicture}[inner sep=0]
            \node [label={[label distance=0.3cm,text depth=-1ex,rotate=90]right: \textcolor{black}{ {Ours}}}] at (0,8.7) {};
            \end{tikzpicture}
			\includegraphics[width=\cimwid\linewidth]{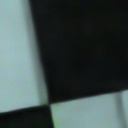}
			\includegraphics[width=\cimwid\linewidth]{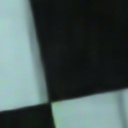}
			\includegraphics[width=\cimwid\linewidth]{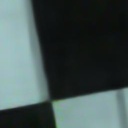}
			\includegraphics[width=\cimwid\linewidth]{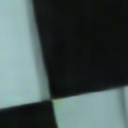}
			\includegraphics[width=\cimwid\linewidth]{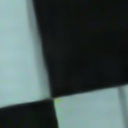}
			\includegraphics[width=\cimwid\linewidth]{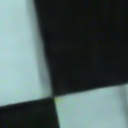}
			\includegraphics[width=\cimwid\linewidth]{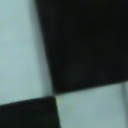}
    \end{minipage}
    \vspace{2em}
	% \caption{Qualitative comparison for the sequence reconstruction  on the RSB dataset. (d),(e),(f) denote ``LEVS"\cite{levs}, ``Moiton-ETR" \cite{motion_etr} and ``BiT" \cite{bit} respectively. }
	% \label{fig:compare_sequence_1}
     \end{minipage}
    \\
    \begin{minipage}{\textwidth}
    	\centering
        \begin{minipage}[t][3.9cm][t]{\rswidth\textwidth}
        \vspace{0pt}
        \centering
              \begin{tikzpicture}[spy using outlines={rectangle,green,magnification=\ssmag,size=\ssizz},inner sep=0]
        				\node {\includegraphics[width=\linewidth]{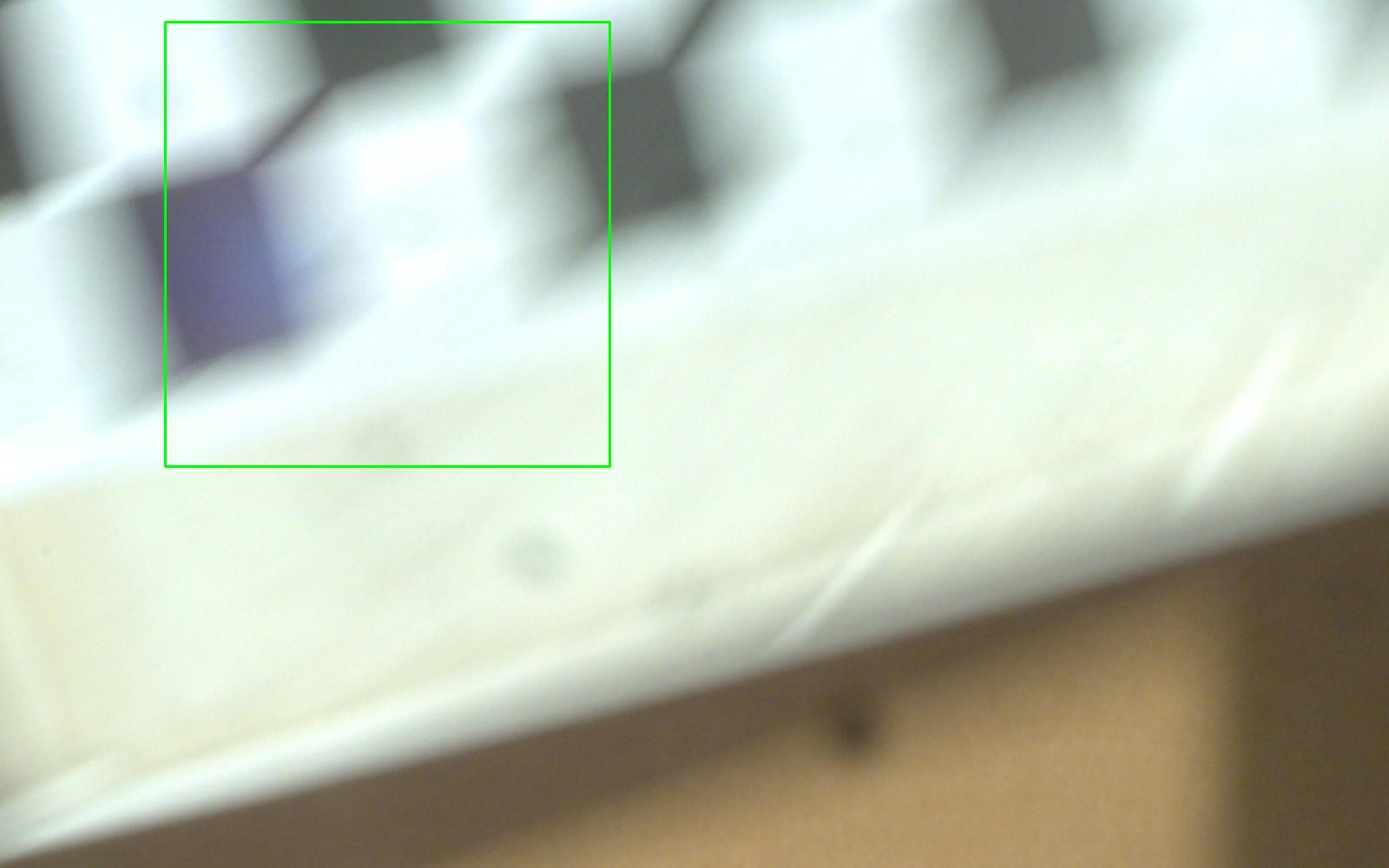}};
        				% \spy on \smallboxloc in node [left] at \largeboxloc;
        				\end{tikzpicture}
            
            Blur\vspace{0.3em}
            % \vspace{0.8em}
                    % RS frame \vspace{-0.3em}
             \vfill
			\begin{tikzpicture}[spy using outlines={rectangle,green,magnification=\ssmag,size=\ssizz},inner sep=0]
				\node {\includegraphics[width=\linewidth]{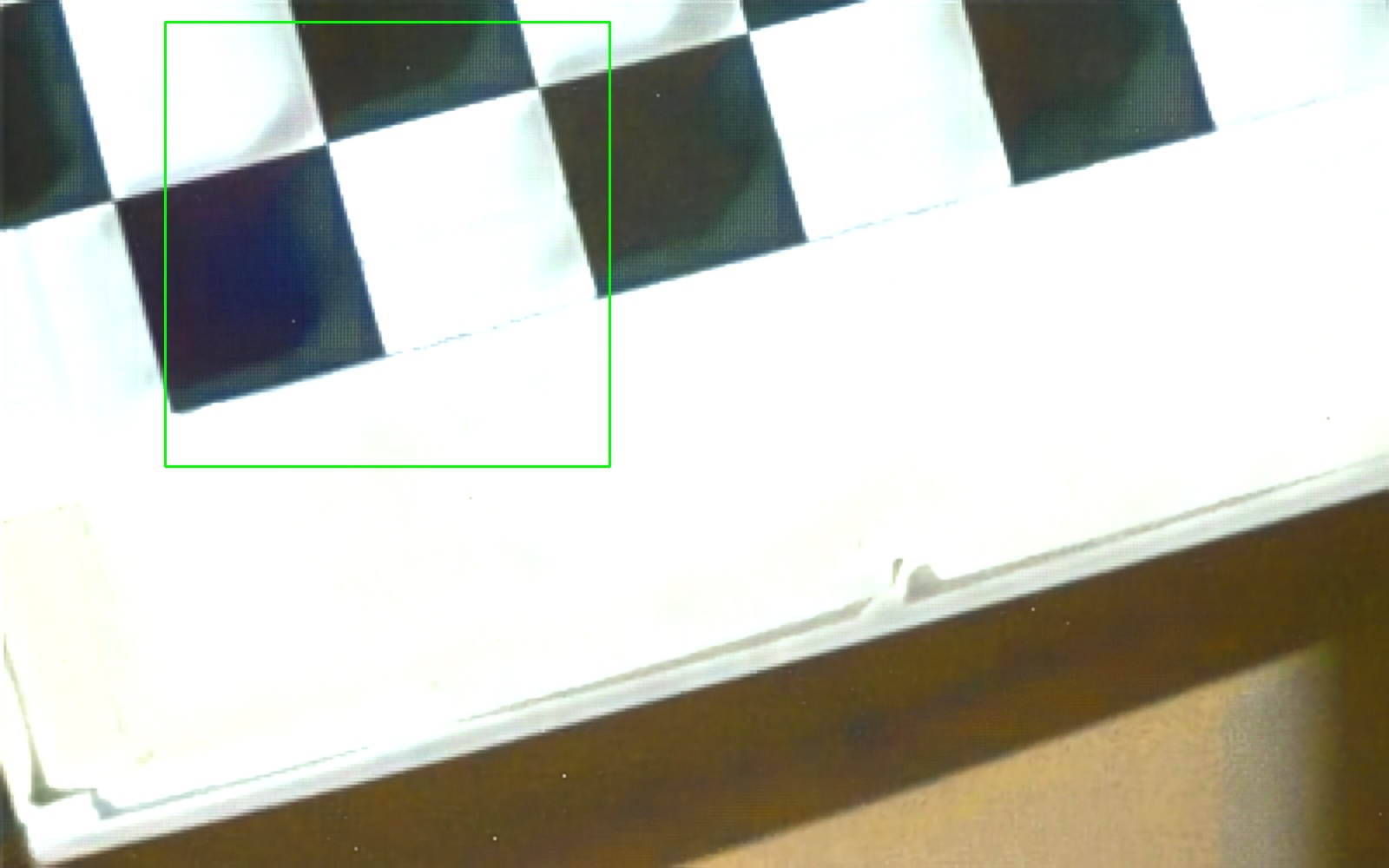}};
				% \spy on \smallboxloc in node [left] at \largeboxloc;
				\end{tikzpicture}
            
           SpkDeblurNet \cite{spike_deblur}\vspace{0.3em}
           \begin{tikzpicture}[spy using outlines={rectangle,green,magnification=\ssmag,size=\ssizz},inner sep=0]
        				\node {\includegraphics[width=\linewidth]{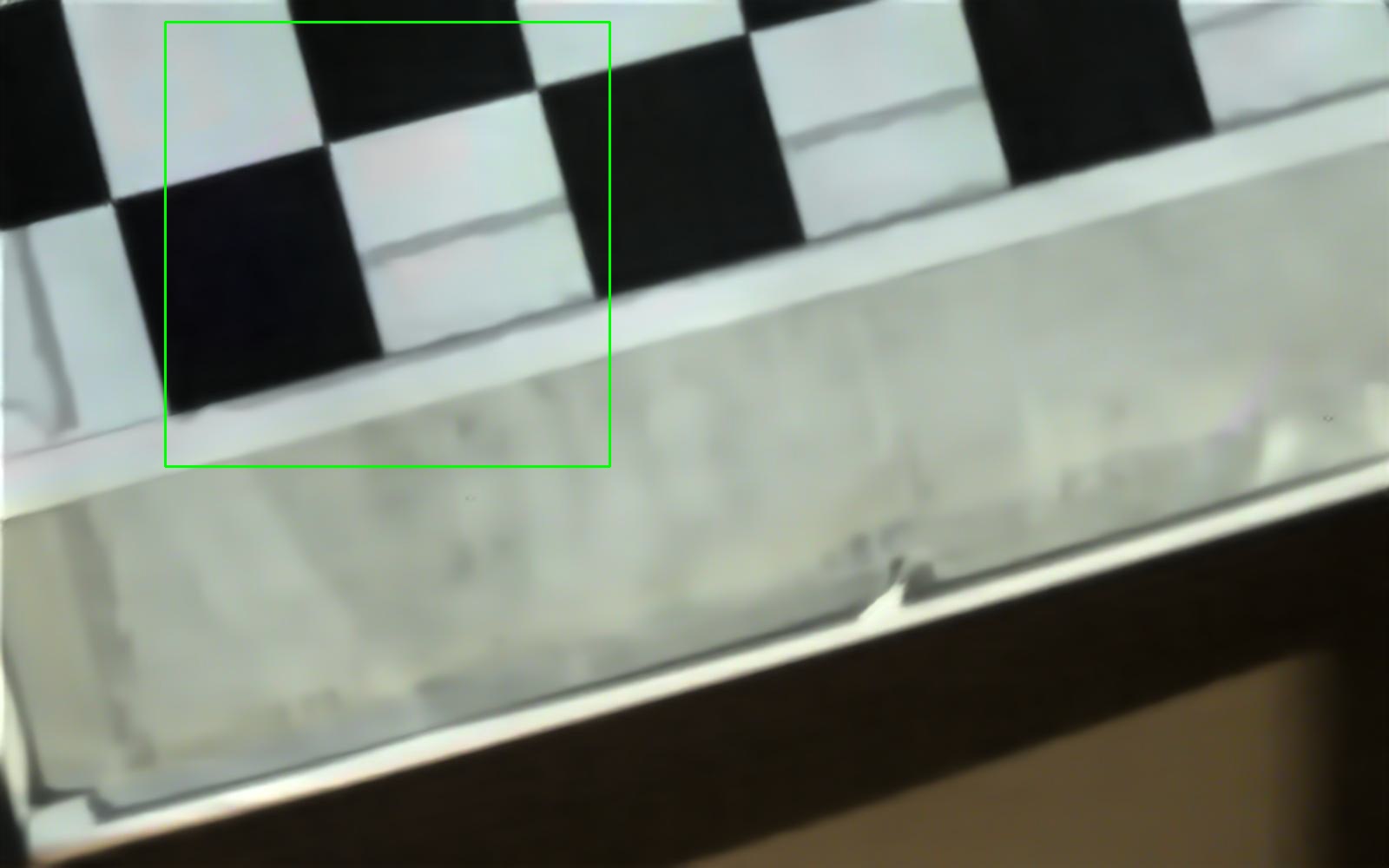}};
        				% \spy on \smallboxloc in node [left] at \largeboxloc;
        				\end{tikzpicture}
            Ours \vspace{0.3em}
    	\end{minipage}% 
% \hspace*{-1mm}
% \hspace*{-5mm}
\begin{minipage}[t]{\seqwidth\textwidth}
            \vspace{0pt}
    		\centering
      % trim={左，下，右，上}
      \begin{tikzpicture}[inner sep=0]
            \node [label={[label distance=0.4cm,text depth=0ex,rotate=90]right: \textcolor{black}{ {(a)} }}] at (0,5) {};
            \end{tikzpicture}
			\includegraphics[width=\cimwid\linewidth]{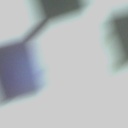}
			\includegraphics[width=\cimwid\linewidth]{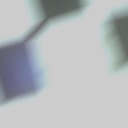}
			\includegraphics[width=\cimwid\linewidth]{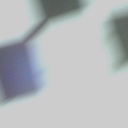}
			\includegraphics[width=\cimwid\linewidth]{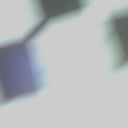}
			\includegraphics[width=\cimwid\linewidth]{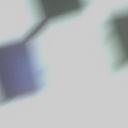}
			\includegraphics[width=\cimwid\linewidth]{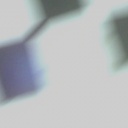}
			\includegraphics[width=\cimwid\linewidth]{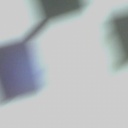}
   \\
     % \\
   \vspace{0.5em}
            \begin{tikzpicture}[inner sep=0]
            \node [label={[label distance=0.4cm,text depth=0,rotate=90]right: \textcolor{black}{ {(b)} }}] at (0,5) {};
            \end{tikzpicture}
			\includegraphics[width=\cimwid\linewidth]{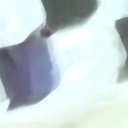}
			\includegraphics[width=\cimwid\linewidth]{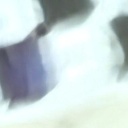}
			\includegraphics[width=\cimwid\linewidth]{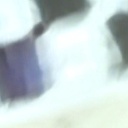}
			\includegraphics[width=\cimwid\linewidth]{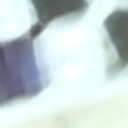}
			\includegraphics[width=\cimwid\linewidth]{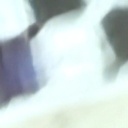}
			\includegraphics[width=\cimwid\linewidth]{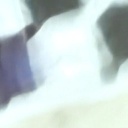}
			\includegraphics[width=\cimwid\linewidth]{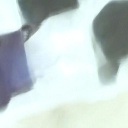}
   \\
   \vspace{0.5em}
            \begin{tikzpicture}[inner sep=0]
            \node [label={[label distance=0.4cm,text depth=-1ex,rotate=90]right: \textcolor{black}{ {(c)}}}] at (0,8.7) {};
            \end{tikzpicture}
			\includegraphics[width=\cimwid\linewidth]{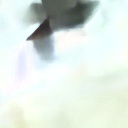}
			\includegraphics[width=\cimwid\linewidth]{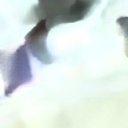}
			\includegraphics[width=\cimwid\linewidth]{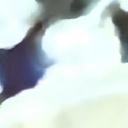}
			\includegraphics[width=\cimwid\linewidth]{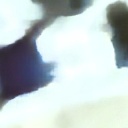}
			\includegraphics[width=\cimwid\linewidth]{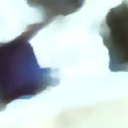}
			\includegraphics[width=\cimwid\linewidth]{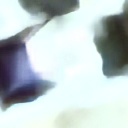}
			\includegraphics[width=\cimwid\linewidth]{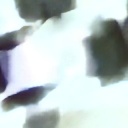}
    \\
   \vspace{0.5em}
            \begin{tikzpicture}[inner sep=0]
            \node [label={[label distance=.2cm,text depth=-1ex,rotate=90]right: \textcolor{black}{ {(d)} }}] at (-20,1) {};
            \end{tikzpicture}
			\includegraphics[width=\cimwid\linewidth]{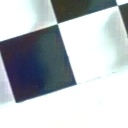}
			\includegraphics[width=\cimwid\linewidth]{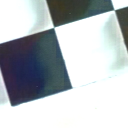}
			\includegraphics[width=\cimwid\linewidth]{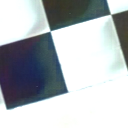}
			\includegraphics[width=\cimwid\linewidth]{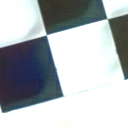}
			\includegraphics[width=\cimwid\linewidth]{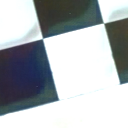}
			\includegraphics[width=\cimwid\linewidth]{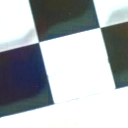}
			\includegraphics[width=\cimwid\linewidth]{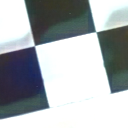}
   \\
   \vspace{0.5em}
            \begin{tikzpicture}[inner sep=0]
            \node [label={[label distance=0.3cm,text depth=-1ex,rotate=90]right: \textcolor{black}{ {Ours}}}] at (0,8.7) {};
            \end{tikzpicture}
			\includegraphics[width=\cimwid\linewidth]{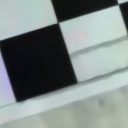}
			\includegraphics[width=\cimwid\linewidth]{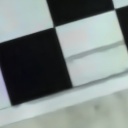}
			\includegraphics[width=\cimwid\linewidth]{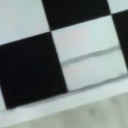}
			\includegraphics[width=\cimwid\linewidth]{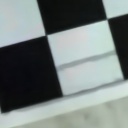}
			\includegraphics[width=\cimwid\linewidth]{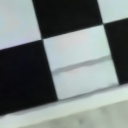}
			\includegraphics[width=\cimwid\linewidth]{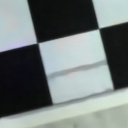}
			\includegraphics[width=\cimwid\linewidth]{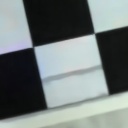}
    \end{minipage}
    \end{minipage}
    \vspace{2em}
    \caption{Qualitative comparison for the sequence reconstruction on the RSB dataset. (a),(b),(c),(d) denote results of  ``LEVS", ``Moiton-ETR" , ``BiT" and ``SpkDeblurNet" respectively. The upper panel depicts a real-world scene with a lower spike density than the simulation, whereas the lower image exhibits a higher spike density. }
	\label{sup_fig:compare_sequence_together}
 \vspace{-1em}
\end{figure*}

Moreover, as depicted in \cref{sup_fig:compare_sequence_together} and \cref{sup_fig:animations}, S-SDM exhibits outstanding performance in precisely retrieving luminance information and texture details while ensuring the motion consistency of reconstructed sequences. In contrast, the video reconstruction of the BiT suffers from poor image quality and shows inadequate sequence continuity in the restored sequence. Besides, SpkDeblurNet encounters issues with color distortion, brightness inconsistency, and inaccurate texture restoration owing to the domain gap between synthetic and real-world datasets. These observations further highlight the superior performance of our S-SDM in real-world scenarios.

\begin{figure}[t]
    \centering
    \includegraphics[width=0.65\linewidth]{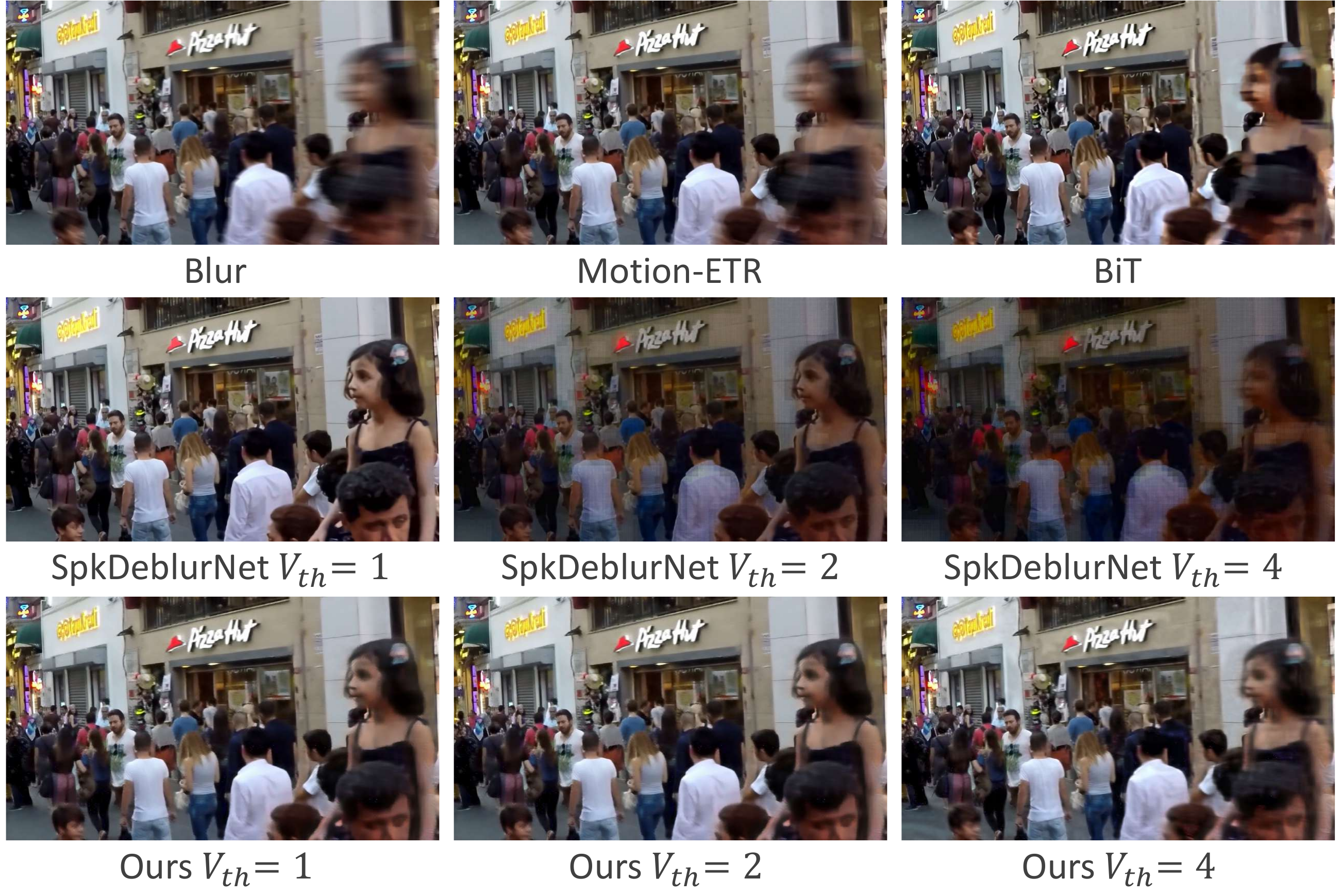}
    \caption{Comparison of our S-SDM against other methods on GOPRO with $V_{th} = 1,2,4$.}
    \label{sup_fig:gopro1}
\end{figure}

\begin{figure}[t]
    \centering
    \includegraphics[width=0.65\linewidth]{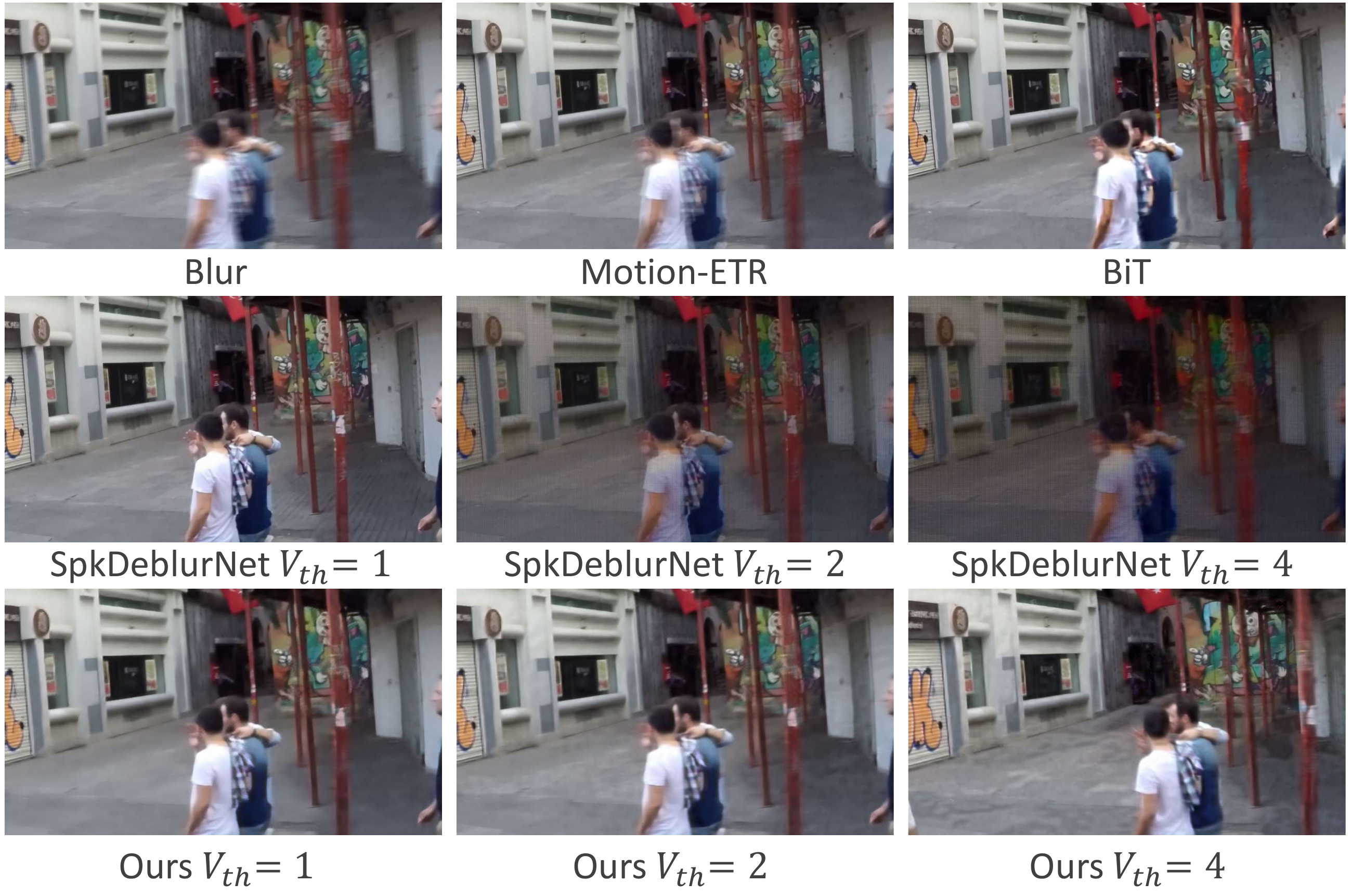}
    \caption{Comparison of our S-SDM against other methods on GOPRO with $V_{th} = 1,2,4$.}
    \label{sup_fig:gopro2}
    \vspace{-1em}
\end{figure}

\textbf{Comparison on the GOPRO dataset.} We provide additional visual comparison of our method against other SOTA methods on the GOPRO dataset as shown in \cref{sup_fig:gopro1,sup_fig:gopro2}.  


% \begin{figure}[t]
%     \centering
%     \includegraphics[width=0.8\linewidth]{imgs/sup_gopro3_compressed.pdf}
%     \caption{Comparison of our S-SDM against other methods on GOPRO  with $V_{th} = 1,2,4$.}
%     \label{sup_fig:gopro3}
% \end{figure}

% \begin{figure}[t]
%     \centering
%     \includegraphics[width=0.8\linewidth]{imgs/sup_gopro4_compressed.pdf}
%     \caption{Comparison of our S-SDM against other methods on GOPRO  with $V_{th} = 1,2,4$.}
%     \label{sup_fig:gopro4}
%     \vspace{-1em}
% \end{figure}

\subsubsection{Implementation}
To augment the dataset and accelerate the training process, we randomly crop  $512 \times 512$ image from each blurry frame, along with the $128 \times 128$ spike stream. We use PyTorch to build and train our S-SDM using an NVIDIA GeForce GTX 4090 GPU and AMD EPYC 7742 64-Core Processor. The training of our LDN consumes about 4 hours on the GOPRO. During the testing phase, we feed the entire image and the spike stream into the network to assess performance.   

We complete the training of BSN on the GOPRO dataset, employing an initial learning rate of $3e^{-4}$ and spanning 1000 epochs. The training uses the Adam optimizer with a cosine scheduler and sets the batch size to 8 for each epoch. Adopting the same settings as BSN, EDSR is trained on the blur-spike paired data for 70 epochs. Subsequently, LDN undergoes the training of 100 epochs with the learning rate adjusted to $1e^{-3}$. 

\begin{figure}[t]
    \centering
    \includegraphics[width=0.9\linewidth]{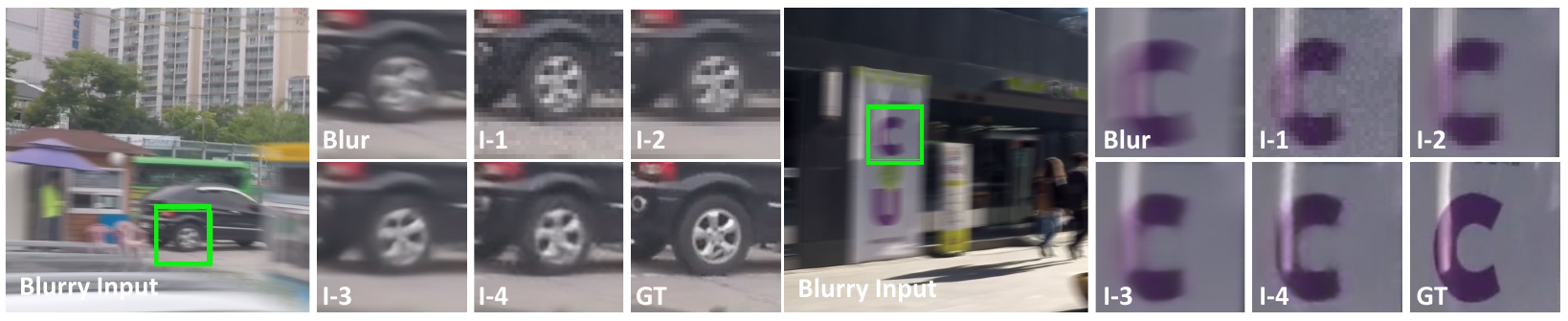}
    \vspace{-1em}
    \caption{Ablation study for evaluating modules of S-SDM on the GOPRO dataset. Experiments corresponding to the ID can be viewed through \cref{tab:ablation1}.}
    \label{sub_fig:ablation1}
    \vspace{-1em}
\end{figure}

\begin{figure}[t]
    \centering
    \includegraphics[width=0.9\linewidth]{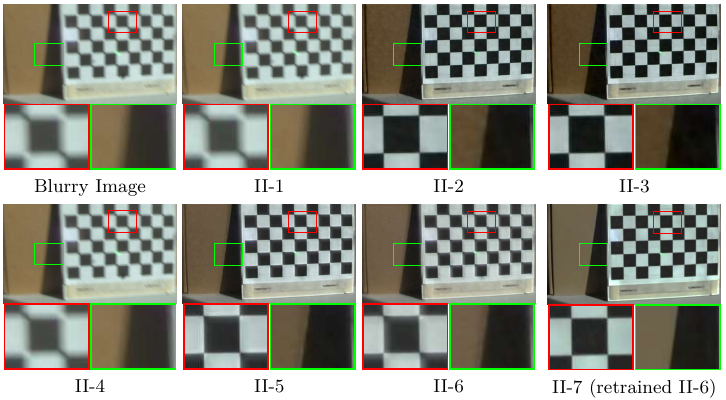}
    \vspace{-1em}
    \caption{Ablation study for evaluating the effectiveness of our distillation learning framework on the RSB dataset. The details corresponding to the experiment ID can be viewed through the \cref{tab:ablation2}.}
    \label{sub_fig:ablation2}
    \vspace{-1em}
\end{figure}

\subsubsection{Ablation study}
We provide additional ablation visualizations to demonstrate the effectiveness of our designed modules and the distillation framework as shown in \cref{sub_fig:ablation1,sub_fig:ablation2}.

\subsection{Additional Figure}\label{sec_sup:figures}

\begin{figure}[h]
    \centering
    \includegraphics[width=0.57\linewidth]{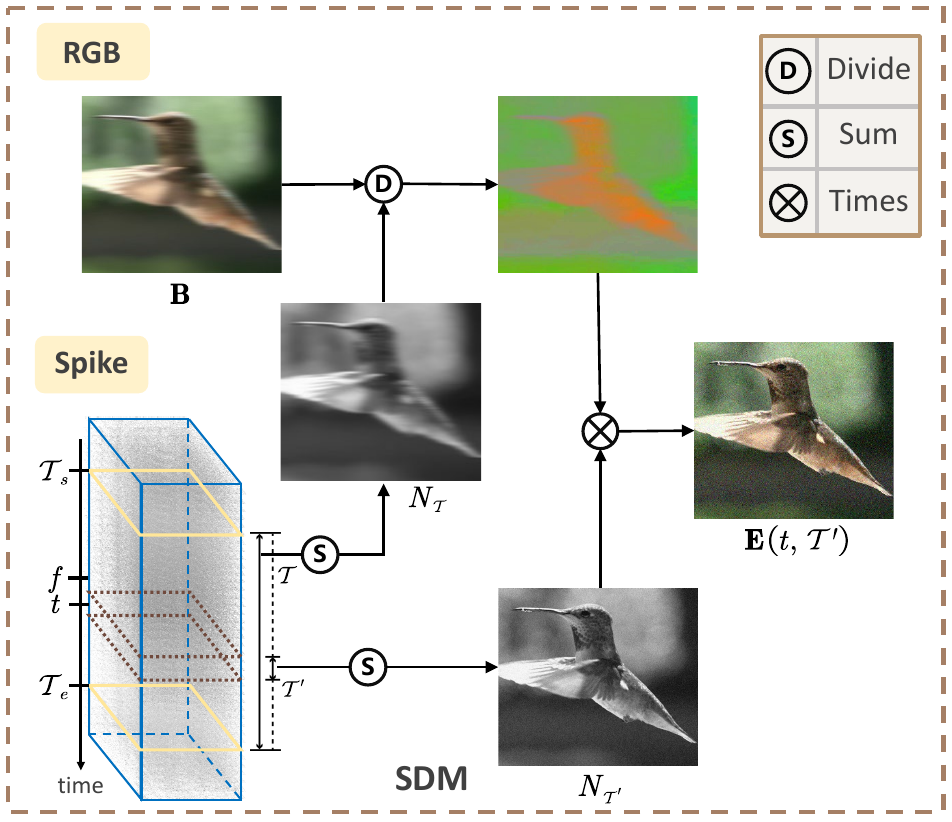}
        \vspace{-1em}
    \caption{The schematic diagram of our designed SDM.}
    \label{sup_fig:s_sdm}
\end{figure}

\newpage
\newpage
\section*{NeurIPS Paper Checklist}

\begin{enumerate}

\item {\bf Claims}
    \item[] Question: Do the main claims made in the abstract and introduction accurately reflect the paper's contributions and scope?
    \item[] Answer: \answerYes{} % Replace by \answerYes{}, \answerNo{}, or \answerNA{}.
    \item[] Justification: See lines 72-77.

\item {\bf Limitations}
    \item[] Question: Does the paper discuss the limitations of the work performed by the authors?
    \item[] Answer: \answerYes{} % Replace by \answerYes{}, \answerNo{}, or \answerNA{}.
    \item[] Justification: See lines 298-299.

\item {\bf Theory Assumptions and Proofs}
    \item[] Question: For each theoretical result, does the paper provide the full set of assumptions and a complete (and correct) proof?
    \item[] Answer: \answerYes{} % Replace by \answerYes{}, \answerNo{}, or \answerNA{}.
    \item[] Justification: This paper provided the theoretical analysis on the spike-guided motion deblurring task.

    \item {\bf Experimental Result Reproducibility}
    \item[] Question: Does the paper fully disclose all the information needed to reproduce the main experimental results of the paper to the extent that it affects the main claims and/or conclusions of the paper (regardless of whether the code and data are provided or not)?
    \item[] Answer: \answerYes{} % Replace by \answerYes{}, \answerNo{}, or \answerNA{}.
    \item[] Justification: Experiments are reproduceable.

\item {\bf Open access to data and code}
    \item[] Question: Does the paper provide open access to the data and code, with sufficient instructions to faithfully reproduce the main experimental results, as described in supplemental material?
    \item[] Answer: \answerYes{} % Replace by \answerYes{}, \answerNo{}, or \answerNA{}.
    \item[] Justification: We have released the code.

\item {\bf Experimental Setting/Details}
    \item[] Question: Does the paper specify all the training and test details (e.g., data splits, hyperparameters, how they were chosen, type of optimizer, etc.) necessary to understand the results?
    \item[] Answer: \answerYes{} % Replace by \answerYes{}, \answerNo{}, or \answerNA{}.
    \item[] Justification: Yes, they are included in the supplementary materials.

\item {\bf Experiment Statistical Significance}
    \item[] Question: Does the paper report error bars suitably and correctly defined or other appropriate information about the statistical significance of the experiments?
    \item[] Answer: \answerNo{} % Replace by \answerYes{}, \answerNo{}, or \answerNA{}.
    \item[] Justification: We don't report it.

\item {\bf Experiments Compute Resources}
    \item[] Question: For each experiment, does the paper provide sufficient information on the computer resources (type of compute workers, memory, time of execution) needed to reproduce the experiments?
    \item[] Answer: \answerYes{} % Replace by \answerYes{}, \answerNo{}, or \answerNA{}.
    \item[] Justification: Yes, they are included in the supplementary materials.
    
\item {\bf Code Of Ethics}
    \item[] Question: Does the research conducted in the paper conform, in every respect, with the NeurIPS Code of Ethics \url{https://neurips.cc/public/EthicsGuidelines}?
    \item[] Answer: \answerYes{} % Replace by \answerYes{}, \answerNo{}, or \answerNA{}.
    \item[] Justification: Yes, we follow the NeurIPS Code of Ethics.

\item {\bf Broader Impacts}
    \item[] Question: Does the paper discuss both potential positive societal impacts and negative societal impacts of the work performed?
    \item[] Answer: \answerNA{} % Replace by \answerYes{}, \answerNo{}, or \answerNA{}.
    \item[] Justification: \answerNA{}
    
\item {\bf Safeguards}
    \item[] Question: Does the paper describe safeguards that have been put in place for responsible release of data or models that have a high risk for misuse (e.g., pretrained language models, image generators, or scraped datasets)?
    \item[] Answer: \answerNA{} % Replace by \answerYes{}, \answerNo{}, or \answerNA{}.
    \item[] Justification: \answerNA{}

\item {\bf Licenses for existing assets}
    \item[] Question: Are the creators or original owners of assets (e.g., code, data, models), used in the paper, properly credited and are the license and terms of use explicitly mentioned and properly respected?
    \item[] Answer: \answerYes{} % Replace by \answerYes{}, \answerNo{}, or \answerNA{}.
    \item[] Justification: Yes, they are correctly used.

\item {\bf New Assets}
    \item[] Question: Are new assets introduced in the paper well documented and is the documentation provided alongside the assets?
    \item[] Answer: \answerNA{} % Replace by \answerYes{}, \answerNo{}, or \answerNA{}.
    \item[] Justification: \answerNA{}

\item {\bf Crowdsourcing and Research with Human Subjects}
    \item[] Question: For crowdsourcing experiments and research with human subjects, does the paper include the full text of instructions given to participants and screenshots, if applicable, as well as details about compensation (if any)? 
    \item[] Answer: \answerNA{} % Replace by \answerYes{}, \answerNo{}, or \answerNA{}.
    \item[] Justification: \answerNA{}

\item {\bf Institutional Review Board (IRB) Approvals or Equivalent for Research with Human Subjects}
    \item[] Question: Does the paper describe potential risks incurred by study participants, whether such risks were disclosed to the subjects, and whether Institutional Review Board (IRB) approvals (or an equivalent approval/review based on the requirements of your country or institution) were obtained?
    \item[] Answer: \answerNA{} % Replace by \answerYes{}, \answerNo{}, or \answerNA{}.
    \item[] Justification: \answerNA{}

\end{enumerate}

\end{document}